\RequirePackage{fix-cm}
\RequirePackage{fixltx2e}
\documentclass[12pt]{article}
\usepackage{fancyhdr}
\usepackage[utf8]{inputenc}
\usepackage[a4paper]{geometry}
\usepackage{float} 
\usepackage{rotfloat}
\usepackage{booktabs}
\usepackage{mathtools}
\usepackage{amsmath}
\usepackage{amsthm}
\usepackage{amssymb}
\RequirePackage{amsfonts}
\usepackage{graphicx}
\usepackage{color}
\usepackage{url}

\renewcommand{\phi}{\varphi}

\newcommand{\grad}{\nabla}

\newcommand{\argmin}{\operatorname*{arg\; min}}

\newcommand{\rank}{\operatorname{rank}}
\newcommand{\diag}{\operatorname{diag}}

\newcommand{\bx}{\boldsymbol{x}}

\newcommand{\by}{\boldsymbol{y}}
\newcommand{\bZ}{\boldsymbol{Z}}

\newcommand{\bz}{\boldsymbol{z}}
\newcommand{\bX}{\boldsymbol{X}}

\newcommand{\bL}{\boldsymbol{L}}

\newcommand{\bP}{\boldsymbol{P}}
\newcommand{\bQ}{\boldsymbol{Q}}
\newcommand{\bD}{\boldsymbol{D}}

\newcommand{\bSigma}{\boldsymbol\Sigma}
\newcommand{\bDelta}{\boldsymbol\Delta}
\newcommand{\bU}{\boldsymbol{U}}

\newcommand{\bV}{\boldsymbol{V}}

\def\reals{\mathbb{R}}
\def\bx{\boldsymbol{x}}

\def\bu{\boldsymbol{u}}
\def\bS{\boldsymbol{S}}
\def\b0{\mathbf{0}}
\def\bP{\boldsymbol{P}}
\def\bQ{\boldsymbol{Q}}
\def\bSigma{\boldsymbol\Sigma}

\def\bDelta{\boldsymbol\Delta}
\def\bU{\boldsymbol{U}}
\def\bR{\boldsymbol{R}}

\def\bv{\boldsymbol{v}}
\def\bA{\boldsymbol{A}}
\def\bAp{\boldsymbol{A}^{\prime}}
\def\bY{\boldsymbol{Y}}
\def\bB{\boldsymbol{B}}
\def\bBp{\boldsymbol{B}^{\prime}}

\def\Col{\mathrm{Col}}

\def\bN{\boldsymbol{N}}

\def\bI{\mathbf{I}}

\def\calM{\mathcal{M}}

\def\bd{\boldsymbol{\delta}}

\makeatletter

\providecommand{\tabularnewline}{\\}
\floatstyle{ruled}
\newfloat{algorithm}{tbp}{loa}
\providecommand{\algorithmname}{{Algorithm}}
\floatname{algorithm}{{\protect\algorithmname}}
 
\theoremstyle{plain}
\newtheorem{lem}{\protect\lemmaname}
\theoremstyle{definition}

\theoremstyle{plain}
\newtheorem{thm}{\protect\theoremname}
\theoremstyle{plain}

\theoremstyle{remark}
\newtheorem{rem}{\protect\remarkname}

\providecommand{\remarkname}{Remark}

\@ifundefined{date}{}{\date{}}

\makeatother
\providecommand{\lemmaname}{Lemma}
\providecommand{\definitionname}{Definition}
\providecommand{\corollaryname}{Corollary} 
\providecommand{\theoremname}{Theorem}
\begin{document}

  \author{Teng Zhang\\
  \multicolumn{1}{p{.9\textwidth}}{\centering\emph{Department of Mathematics, University of Central Florida}\\ teng.zhang@ucf.edu }
  \\
  \\
 Yi Yang\\\multicolumn{1}{p{.9\textwidth}}{\centering\emph{Department of Mathematics and Statistics, McGill University}\\ yi.yang6@mcgill.ca}
  }

\title{\textbf{Robust PCA by Manifold Optimization}}

\maketitle

\begin{abstract} Robust PCA is a widely used statistical procedure to recover a underlying low-rank matrix with grossly corrupted observations. This work considers the problem of robust PCA as a nonconvex optimization problem on the manifold of low-rank matrices, and proposes two algorithms (for two versions of retractions) based on manifold optimization. It is shown that, with a proper designed initialization, the proposed algorithms are guaranteed to converge to the underlying low-rank matrix linearly. Compared with a previous work based on the Burer-Monterio decomposition of low-rank matrices~\cite{DBLP:conf/nips/YiPCC16}, the proposed algorithms reduce the dependence on the conditional number of the underlying low-rank matrix theoretically. Simulations and real data examples confirm the competitive performance of our method.
\end{abstract}
\section{Introduction}

In many data science problems, such as in computer vision~\cite{Epstein95,Ho03}, machine learning~\cite{ASI:ASI1}, and bioinformatics~\cite{Price2006}, the underlying data matrix is assumed to be approximately low-rank. Principal component analysis (PCA) is a standard statistical procedure to recover such underlying low-rank matrix. However, PCA is highly sensitive to outliers in the data, and robust PCA ~\cite{robust_pca09,Chandrasekaran_Sanghavi_Parrilo_Willsky_2009,Clarkson:2013:LRA:2488608.2488620,Frieze:2004:FMA:1039488.1039494,Bhojanapalli:2015:TLA:2722129.2722191,DBLP:conf/nips/YiPCC16,ChenWainwright2015,conf/aistats/GuWL16,DBLP:journals/corr/CherapanamjeriG16,NIPS2014_5430} is hence proposed as a modification to handle grossly corrupted observations. It has been shown to have applications in many fields including background detection~\cite{Li_backgroundsubtraction}, face recognition~\cite{Basri03}, ranking, and collaborative filtering \cite{robust_pca09}. Mathematically, the robust PCA problem is formulated as follows: suppose that given a data matrix $\bY\in\reals^{n_1\times n_2}$ that can be written as the sum of a low-rank matrix $\bL^*$ (signal) and a sparse matrix $\bS^*$ (noise) with only a few nonzero elements, can we recover both components accurately? While there are many algorithms proposed for solving robust PCA, we only review the ones that have the theoretical guarantee on the recovery of underlying low-rank matrix. 

Given the fact that the set of all low-rank matrix is nonconvex, it is generally very difficult to obtain a theoretical guarantee since there is no tractable optimization algorithm for this nonconvex problem. To solve this issue, the works \cite{robust_pca09,Chandrasekaran_Sanghavi_Parrilo_Willsky_2009} consider the convex relaxation of the original problem instead,
\begin{equation}\label{eq:convex}
\min_{\bL,\bS}\|\bL\|_*+\|\bS\|_1,\,\,\text{s.t. $\bY=\bL+\bS$},
\end{equation}
where $\|\bL\|_*$ represents the nuclear norm (i.e., Schatten $1$-norm) of $\bL$, defined by the sum of its singular values and $\|\bS\|_1$ represents the sum of the absolute values of all elements of $\bS$. Since this problem is convex, its global minimizer can be solved efficiently. In addition, it is shown that the global minimizer recovers the correct low-rank matrix when $\bS^*$ has at most $\gamma=O(1/\mu^2r)$ fraction of corrupted non-zero entries, where $r$ is the rank of $\bL^*$ and $\mu$ is the incoherence level of $\bL^*$~\cite{5934412}. If the sparsity of $\bS^*$ is assumed to be random, then \cite{robust_pca09} shows that the algorithm succeeds with high probability, even when the percentage of corruption can be in the order of $O(1)$ while the rank $r=O(\min(n_1,n_2)/\mu\log^2\max(n_1,n_2))$ (this work defines $\mu$ slightly differently, but the value is comparable).

However, the aforementioned algorithms based on convex relaxation have a complexity of $O(n_1n_2\min(n_1,n_2))$ per iteration, which could be prohibitive when $n_1$ and $n_2$ are very large. Alternatively, some other algorithms based on non-convex optimization are proposed. In particular, the work by \cite{6319655} proposes a method based on the projected gradient method. However, it assumes that the sparsity pattern of $\bS^*$ is random, and the algorithm still has the same computational complexity as the convex methods. \cite{NIPS2014_5430} proposes a method based on the alternating projecting, which allows $\gamma\leq \frac{1}{\mu^2 r}$, with the computational complexity of $O(r^2n_1n_2)$ per iteration. \cite{ChenWainwright2015} assumes that $\bL^*$ is positive semidefinite and applies the gradient descent method on the Cholesky decomposition factor of $\bL^*$, but the positive semidefinite assumption is usually not satisfied in practice. \cite{conf/aistats/GuWL16} decomposes $\bL^*$ into the product of two matrices and performs alternating minimization over both matrices. It shows that the algorithm allows $\gamma=O(1/\mu^{2/3}r^{2/3}\min(n_1,n_2))$ and has the complexity of $O(r^2n_1n_2)$ per iteration. \cite{DBLP:conf/nips/YiPCC16} applies a similar decomposition and applies the gradient descent algorithm with a complexity of $O(rn_1n_2)$ per iteration and allows $\gamma=O(1/\kappa^2\mu r^{3/2})$, where $\kappa$ is the conditional number of the underlying low-rank matrix.  \cite{NIPS2014_5430} proposes a method based on alternating projection, which has a complexity of $O(r^2n_1n_2)$ per iteration. They show that the algorithm can still succeed when the corruption level $\gamma=O(1/\mu^2 r)$. There is another line of works that further reduces the complexity of the algorithm by subsampling the entries of the observation matrix $\bY$, including \cite{NIPS2011_4486,7035075,7809181,DBLP:journals/corr/CherapanamjeriG16} and \cite[Algorithm 2]{DBLP:conf/nips/YiPCC16}, we will discuss it later in Section~\ref{sec:partial}. 

The common idea shared by \cite{conf/aistats/GuWL16} and \cite{DBLP:conf/nips/YiPCC16} is as follows. Since any low-rank matrix $\bL\in\reals^{n_1\times n_2}$ with rank $r$ can be written as the product of two low-rank matrices by $\bL=\bU\bV^T$ with $\bU\in\reals^{n_1\times r}$ and $\bV\in\reals^{n_2\times r}$, we can optimize the pair $(\bU,\bV)$ instead of $\bL$, and a smaller computational cost is expected since $(\bU,\bV)$ has $(n_1+n_2)r$ parameters, which is smaller than $n_1n_2$, the number of parameters in $\bL$. In fact, such a re-parametrization technique has a long history~\cite{ruhe1974numerical}, and has been  popularized by Burer and Monteiro~\cite{Burer2003,Burer2005} for solving semi-definite programs (SDPs). The same idea has been used in other low-rank matrix estimation problems such as dictionary learning~\cite{7755794}, phase synchronization~\cite{doi:10.1137/16M105808X}, community detection~\cite{pmlr-v49-bandeira16}, matrix completion~\cite{Jain:2013:LMC:2488608.2488693}, recovering matrix from linear measurements~\cite{DBLP:conf/icml/TuBSSR16}, and even general problems~\cite{ChenWainwright2015,pmlr-v54-wang17b,Park2016,pmlr-v54-wang17b,pmlr-v54-park17a}. In addition, the property of associated stochastic gradient descent algorithm is studied in~\cite{DeSa:2015:GCS:3045118.3045366}, 

The main contribution of this work is a new algorithm for solving the robust PCA, based on the gradient descent algorithm on the manifold of low-rank matrices, with a theoretical guarantee on the exact recovery of the underlying low-rank matrix. Compared with \cite{DBLP:conf/nips/YiPCC16}, the proposed algorithm   utilizes the tool of manifold optimization, which leads to a simpler and more naturally structured algorithm with a stronger theoretical guarantee. In particular, with a proper initialization, our method can still succeeds with $\gamma=O(1/\kappa\mu r^{3/2})$, which means that it can tolerate more corruption than \cite{DBLP:conf/nips/YiPCC16} by a factor of $\kappa$. In addition, the theoretical convergence rate is also faster than \cite{DBLP:conf/nips/YiPCC16} by a factor of $\kappa$. Simulations also verified the advantage of the proposed algorithm over \cite{DBLP:conf/nips/YiPCC16}. Considering the popularity of Burer-Monteiro decomposition, we expect that manifold optimization could be applied to other low-rank matrix estimation problems.

The paper is organized as follows. We review the background of manifold optimization and the manifold of low-rank matrices in Section~\ref{sec:review_algo}. In Section \ref{sec:alg}, we present our gradient descent algorithms on the manifold. We analyze their theoretical properties and compare them with previous algorithms in Section~\ref{sec:theory}.  In Section~\ref{sec:simu}, we conduct simulations and real data analysis on the  \texttt{Shoppingmall} dataset to show that our algorithms have superior performances to the original gradient descent on the factorized space. A discussion about the proposed algorithms is then presented in Section \ref{sec:discussion}, followed by the proofs of the results in Appendix. 

\section{Algorithm}\label{sec:review_algo}
\subsection{Optimization on manifold}\label{sec:background}
The purpose of this section is to review the framework of  the gradient descent method on manifolds. It summarizes mostly the framework used in \cite{Vandereycken2013,Shalit:2012:OLE:2503308.2188399,absil2009optimization}, and we refer  readers to these work for more details.

Given a smooth manifold $\calM\subset\reals^n$ and a differentiable function $f: \calM\rightarrow \reals$, the procedure of the gradient descent algorithm for solving $\min_{x\in\calM} f(x)$ is as follows:
\begin{description}

\item[Step 1.] Consider $f(x)$ as a function from $\reals^n$ to $\reals$ and calculate the Euclidean gradient $\grad f(x)$.

\item[Step 2.] Calculate its Riemannian gradient, which is the direction of steepest ascent of $f(x)$ among all directions in the \textit{tangent space} $T_x\calM$. This direction is given by $P_{T_x\calM}\grad f(x)$, where $P_{T_x\calM}$ is the projection operator to the tangent space $T_x\calM$.

\item[Step 3.]  Define a \textit{retraction} $R_x$ that maps the tangent space back to the manifold, i.e. $R_x: T_x\calM\rightarrow \calM$, where $R_x$ needs to satisfy the conditions in \cite[Definition 2.2]{Vandereycken2013}. In particular, $R_x(0)=x$, $R_x(y)=x+y+o(\|y\|^2)$ as $y\rightarrow 0$, and $R_x$ needs to be smooth. Then the update of the gradient descent algorithm $x^+$ is defined by
\begin{equation}\label{eq:gradient_manifold}
x^+=R_x(-\eta P_{T_x\calM}\grad f(x)),
\end{equation}
where $\eta$ is the step size. 
\end{description}

\begin{figure}
\begin{center}
\includegraphics[width=0.6\textwidth]{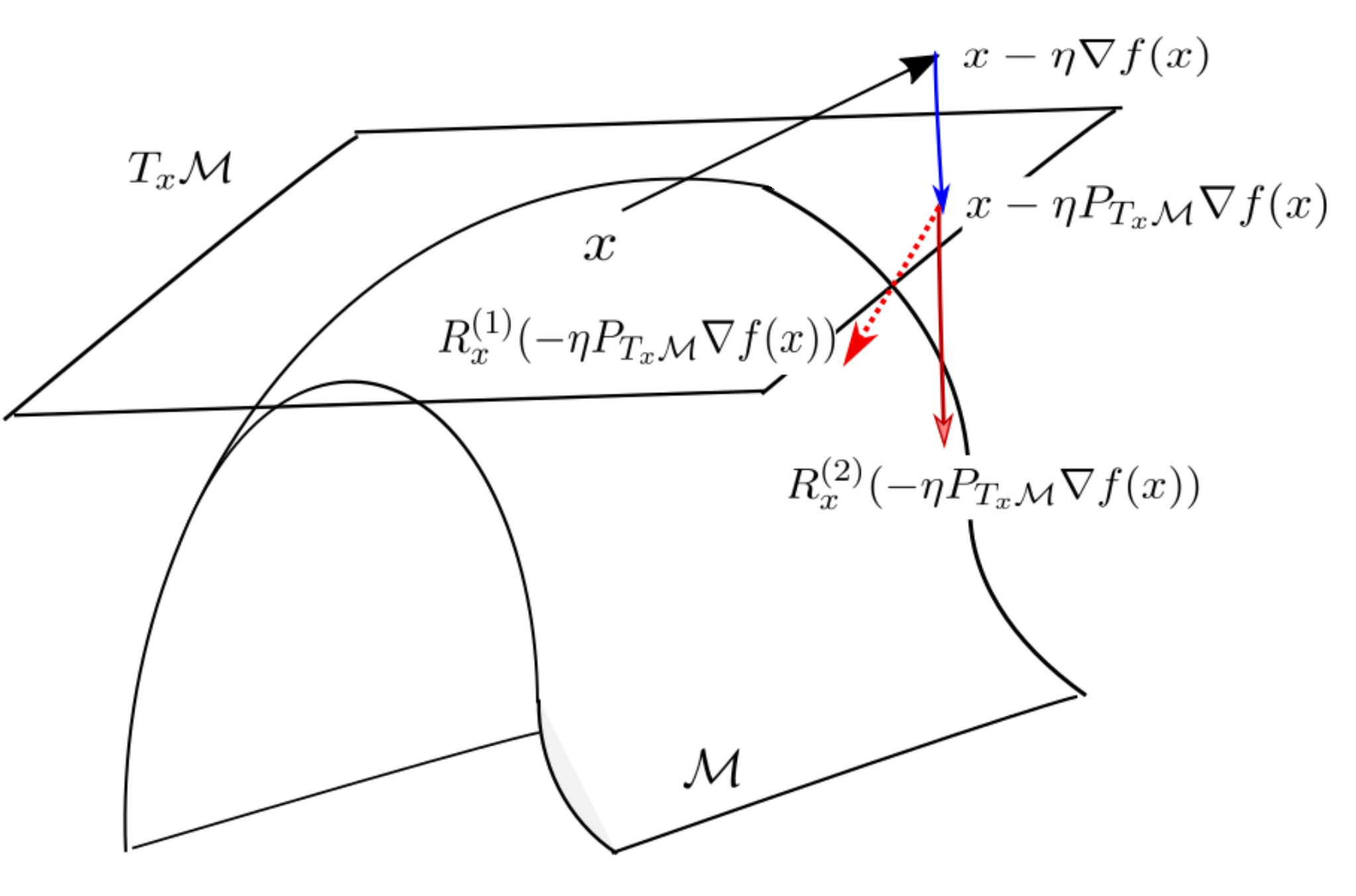}
\end{center}
\caption{The visualization of gradient descent algorithm on the manifold $\mathcal{M}$. The black solid line is the Euclidean gradient. The blue solid line is the projection of the Euclidean gradient to the tangent space. The red solid line represents the orthographic retraction, while the red dashed line represents the projective retraction.}\label{fig:visualization}
\end{figure}

Note that the definition of retraction is not unique. In Figure~\ref{fig:visualization}, we visualize the gradient descent method on the manifold $\calM$ with two different kinds of retractions (orthographic and projective). We will discuss the details of those two retractions in Section \ref{sec:geometry}.

\subsection{The geometry of the manifold of low-rank matrices}\label{sec:geometry}
Now we apply the above gradient descent algorithm to the manifold of the low-rank matrices. Let $\calM$ be the manifold of all $\reals^{n_1\times n_2}$ matrices with rank $r$ and denote $\bX\in\calM$ a specific matrix . The tangent space $T_{\bX}\calM$ and the retraction $R_{\bX}$ of the manifold of the low-rank matrices have been well-studied~\cite{Absil2015}. The tangent space $T_{\bX}\calM$ can be defined by $T_{\bX}\calM=\{\bA\bX+\bX\bB\}$ where $\bA\in\reals^{n_1\times n_1}$ and  $\bB\in\reals^{n_2\times n_2}$ according to \cite{Absil2015}. The explicit formula for the projection $P_{T_{\bX}\calM}\grad f$ is given in \cite[(9)]{Absil2015}. Assume $\bD=\grad f(\bX)$. Denote the SVD decomposition of $\bX$ as $\bX=\bU\bSigma\bV^{T}$, then 
\begin{equation}\label{eq:projection}
P_{T_{\bX}\calM}(\bD)=\bU\bU^{T}\bD+ \bD\bV\bV^{T}-\bU\bU^{T}\bD\bV\bV^{T}.
\end{equation}
The formula \eqref{eq:projection} can be verified as follows. Note that $T_{\bX}\calM$ can be equivalently defined by $\{\bA\bX+\bX\bB\}=\{\bAp\bV\bV^{T}+\bU\bU^{T}\bBp\}$ for any matrices $\bAp\in\reals^{n_1\times n_1}$,  $\bBp\in\reals^{n_2\times n_2}$, such that $\bAp\bV=\bA\bU\bSigma$ and $\bU^{T}\bBp=\bSigma\bV^{T}\bB$, thus $P_{T_{\bX}\calM}(\bD)\in T_{\bX}\calM$. Furthermore, let $\langle\cdot\rangle$ be the Frobenius inner product of two matrices, then 
\begin{align*}
&\langle \bD-P_{T_{\bX}\calM}(\bD),  \bAp\bV\bV^{T}\rangle=\langle (\bI-\bU\bU^{T})\bD(\bI-\bV\bV^{T}),  \bAp\bV\bV^{T}\rangle\\=&\langle (\bI-\bU\bU^{T})\bD(\bI-\bV\bV^{T})\bV\bV^{T},  \bAp\rangle=\langle \mathbf{0},\bAp\rangle=0
\end{align*}
and similarly $\langle \bD-P_{T_{\bX}\calM}(\bD),  \bU\bU^{T}\bBp\rangle=0$. As a result, $\langle \bD-P_{T_{\bX}\calM}(\bD),   \bAp\bV\bV^{T}+\bU\bU^{T}\bBp\rangle=0$ for all $\bAp\in\reals^{n_1\times n_1}$ and  $\bBp\in\reals^{n_2\times n_2}$, which verifies the formula \eqref{eq:projection}.

There are various ways of defining retractions for the manifold of low-rank matrices, and we refer the reader to~\cite{Absil2015} for more details. In this work, we consider only two types of retractions. One is called the \textit{projective} retraction \cite{Shalit:2012:OLE:2503308.2188399,Vandereycken2013}, defined as the nearest low-rank matrix to $\bX+\bd$ in terms of Frobenius norm: 
\begin{equation}\label{eq:projective}
R_{\bX}^{(1)}(\bd)=\argmin_{\bZ\in\calM }\|\bX+\bd-\bZ\|_F,
\end{equation}
for $\bd\in T_{\bX}\calM$ and the solution is given by $\sum_{i=1}^r\sigma_i\bu_i\bv_i^T$, where $\sigma_i,\bu_i,\bv_i$ are the ordered singular values and vectors of $\bX+\bd$ respectively. In order to further improve computation efficiency,  we also consider the \textit{orthographic} retraction \cite{Absil2015}. Denoted by $R^{(2)}_{\bX}(\bd)$, it is the nearest low-rank matrix to $\bX+\bd$ such that 
\begin{align}\label{eq:projective}
&R_{\bX}^{(2)}(\bd)=\argmin_{\bZ\in\calM }\|\bX+\bd-\bZ\|_F,\nonumber\\
&\text{s.t. }\langle R^{(2)}_{\bX}(\bd)-\bX-\bd, \bZ\rangle=0\,\,\text{for all $\bZ\in  T_{\bX}\calM$}. 
\end{align} 
\cite[Section 3.2]{Absil2015} gives an explicit formula for the orthographic retraction, 
\begin{equation}\label{eq:ortho_retraction}
R^{(2)}_{\bX}(\bd)=(\bX+\bd)\bV[\bU^{T}(\bX+\bd)\bV]^{-1}\bU^{T}(\bX+\bd).
\end{equation}
Later we will show that the solution has a simple explicit formula and there is no need to calculate singular value decomposition as the projective retraction.

\section{Proposed algorithms}\label{sec:alg}
To  recover $\bL^{*}$, we solve the following optimization problem:
\begin{equation}\label{eq:objective}
{  \argmin_{\rank(\bL)=r}f(\bL), \,\,\text{where $f(\bL)=\frac{1}{2}\|F(\bL-\bY)\|_F^2$},}
\end{equation}
where $F: \reals^{n_1\times n_2}\rightarrow\reals^{n_1\times n_2}$ is a hard thresholding procedure defined by
\begin{equation}\label{eq:threshold}
F_{ij}(\bA)=\begin{cases}0, &\text{if $|\bA_{ij}|> |\bA_{i,\cdot}^{[\gamma ]}|$ and $|\bA_{ij}|> |\bA_{\cdot,j}^{[\gamma ]}|$}\\
\bA_{ij}, &\text{otherwise}.\end{cases}
\end{equation}
Here $\bA_{i,\cdot}$ represents the $i$-th row of $\bA$, and $\bA_{\cdot,j}$ represents the $j$-th column of $\bA$. $\bA_{i,\cdot}^{[\gamma]}$ and $\bA_{\cdot,j}^{[\gamma]}$ represent the $\gamma$-th percentile of the absolute values of the entries of the $i$-th row and the $j$-th column respectively.
In other words, those elements that are simultaneously among the largest $\gamma$-fraction entries in terms of absolute values in the corresponding row and column of $\bA$ are removed. The threshold $\gamma$ is set by users. If some entries of $\bA_{i,\cdot}$ or $\bA_{\cdot,j}$ have the elements with identical absolute values, the ties can be broken down arbitrarily. 

By applying \eqref{eq:gradient_manifold}, the iterative algorithm for solving \eqref{eq:objective} can be written by
\begin{equation}\label{eq:gradient}
\bL^{(k+1)}=R_{\bL^{(k)}}(-\eta P_{T_{\bL^{(k)}}} \grad f(\bL^{(k)})).
\end{equation}
In the following we provide the explicit formulas for the gradient $\grad f$, projection $P_{T_{\bL^{(k)}}}$ and retraction operations $R_{\bL^{(k)}}$  in \eqref{eq:gradient}.

To find $\grad f$, we define the operator  $S: \reals^{n_1\times n_2}\rightarrow\reals^{n_1\times n_2}$: 
\[
\bS(\bA)=\begin{cases} 0,\,\,\,\, \text{if the first case in \eqref{eq:threshold} is satisfied,}\\
1,\,\,\,\, \text{if the second case in \eqref{eq:threshold} is satisfied.}\end{cases}
\]
Then if the absolute values of all entries of $\bA$ are different, the sparsity pattern does not change under a small perturbation, i.e., $\bS(\bA)=\bS(\bA+\bDelta).$ Then by definition of $f(\cdot)$,
\begin{align*}
&f(\bL+\bDelta)-f(\bL)=\frac{1}{2}\|\bS(\bL-\bY+\bDelta)\circ (\bL-\bY+\bDelta)\|_F^2-\frac{1}{2}\|\bS(\bL-\bY)\circ (\bL-\bY)\|_F^2\\
=&\frac{1}{2}\|\bS(\bL-\bY)\circ (\bL-\bY+\bDelta)\|_F^2-\frac{1}{2}\|\bS(\bL-\bY)\circ (\bL-\bY)\|_F^2\\
=&\langle\bS(\bL-\bY)\circ (\bL-\bY),\bDelta\rangle_{F}+O(\|\bDelta\|_F^2),
\end{align*}
where $\circ$ represents the Hadamard product, i.e., the elementwise product  between matrices. This implies
\begin{equation}\label{eq:derivative}
\grad f(\bL)=\bS(\bL-\bY)\circ (\bL-\bY)=F(\bL-\bY).
\end{equation}
Therefore, the gradient descent algorithm with projective retraction can be written as follows, with $P_{T_{\bL^{(k)}}}$ defined later in \eqref{eq:projection}: 
\begin{equation}\label{eq:alg_projective}
\bL^{(k+1)}\coloneqq \text{rank-$r$ approximation of }\left[\bL^{(k)}-\eta P_{T_{\bL^{(k)}}} F(\bL^{(k)}-\bY)\right],
\end{equation}
and for orthographic retraction,
\begin{equation}\label{eq:alg_orthographic}
\bL^{(k+1)}\coloneqq R^{(2)}_{\bL^{(k)}}\left[\bL^{(k)}-\eta P_{T_{\bL^{(k)}}} F(\bL^{(k)}-\bY)\right].
\end{equation}
Now we get the explicit formula for the projection { $P_{T_{\bL^{(k)}}\calM}\grad f(\bL^{(k)})$. Write $\bD=\grad f(\bL^{(k)})$.} Given the SVD decomposition $\bL^{(k)}=\bU^{(k)}\bSigma^{(k)}\bV^{(k)T}$, we have the projection
 \begin{equation}\label{eq:projection}
 P_{T_{\bL^{(k)}}\calM}(\bD)=\bU^{(k)}\bU^{(k)T}\bD+
  \bD\bV^{(k)}\bV^{(k)T}-\bU^{(k)}\bU^{(k)T}\bD\bV^{(k)}\bV^{(k)T}.
\end{equation}
To compute the projective retraction $R^{(1)}_{\bL^{(k)}}(\bd)$, where $\bd=-\eta P_{T_{\bL^{(k)}}\calM}(\bD)$, we get the singular value decomposition $\bL^{(k)}+\bd=\bar{\bU}\bar{\bSigma}\bar{\bV}^T$. The projective retraction is
 \[
 R^{(1)}_{\bL^{(k)}}(\bd)=\bU^{(k+1)}\bSigma^{(k+1)}\bV^{(k+1)T},
 \]
where $\bU^{(k+1)}\in\reals^{n_1\times r}$ is a matrix consists of the first $r$ columns of $\bar{\bU}$; $\bV^{(k+1)}\in\reals^{n_2\times r}$ is a matrix consists of the first $r$ columns of $\bar{\bV}$; $\bSigma^{(k+1)}\in\reals^{r\times r}$ is the upper left $r\times r$ submatrix of $\bar{\bSigma}$, and $\bL^{(k+1)}=\bU^{(k+1)}\bSigma^{(k+1)}\bV^{(k+1)T}$. The orthographic retraction is
 \begin{equation}\label{eq:ortho_retraction}
 R^{(2)}_{\bL^{(k)}}(\bd)=(\bL^{(k)}+\bd)\bV^{(k)}[\bU^{(k)T}(\bL^{(k)}+\bd)\bV^{(k)}]^{-1}\bU^{(k)T}(\bL^{(k)}+\bd).
 \end{equation}
We remark that for the formula \eqref{eq:alg_orthographic} can be further simplified. Note that
\begin{align}\label{eq:dv}
&P_{T_{\bL^{(k)}}\calM}(\bD)\bV^{(k)}=[\bU^{(k)}\bU^{(k)T}\bD+
 \bD\bV^{(k)}\bV^{(k)T}-\bU^{(k)}\bU^{(k)T}\bD\bV^{(k)}\bV^{(k)T}]\bV^{(k)}\nonumber\\
 &=\bU^{(k)}\bU^{(k)T}\bD\bV^{(k)}+\bD\bV^{(k)}\bV^{(k)T}\bV^{(k)}-\bU^{(k)}\bU^{(k)T}\bD\bV^{(k)}\bV^{(k)T}{ \bV^{(k)}} =\bD\bV^{(k)}
\end{align}
and similarly,
\[
\bU^{(k)T}P_{T_{\bL^{(k)}}\calM}(\bD)=\bU^{(k)T}\bD.
\]
Let $\bD=F(\bL^{(k)}-\bY)$ and $\bd=-\eta P_{T_{\bL^{(k)}}\calM}(\bD)$, by \eqref{eq:derivative} and \eqref{eq:dv}, we have \[\bd\bV^{(k)}=-\eta\bD\bV^{(k)}=-\eta\grad f(\bL^{(k)})\bV^{(k)}=-\eta F(\bL^{(k)}-\bY)\bV^{(k)}\]
and
\[\bU^{(k)}\bd=-\bU^{(k)}\eta\grad f(\bL^{(k)})=-\bU^{(k)}\eta F(\bL^{(k)}-\bY).\]
So the update formula \eqref{eq:alg_orthographic} can be simplified to
\begin{equation}\label{eq:ortho_retraction1}
\bL^{(k+1)}\coloneqq R_{\bL^{(k)}}^{(2)}(\bd)
=(\bL^{(k)}-\eta \bD)\bV^{(k)}[\bU^{(k)T}(\bL^{(k)}-\eta \bD)\bV^{(k)}]^{-1}\bU^{(k)T}(\bL^{(k)}-\eta \bD).
\end{equation}

In addition, it can be shown that $\bU^{(k)}$ and $\bV^{(k)}$ in \eqref{eq:ortho_retraction} and \eqref{eq:ortho_retraction1} can be replaced by $\bU^{(k)}\bA$  and $\bV^{(k)}\bB$ for any nonsingular matrices $\bA,\bB\in\reals^{r\times r}$. That is,  $\bU^{(k)}$ and $\bV^{(k)}$ in \eqref{eq:ortho_retraction} {and \eqref{eq:ortho_retraction1}} can be replaced by any matrices $\bQ\in\reals^{n_1\times r}$ and $\bR\in\reals^{n_2\times r}$ that have the same column spaces as $\bU^{(k)}$ and $\bV^{(k)}$ respectively.

The complete procedures of the implementation are summarized in Algorithm~\ref{alg:gradient} and Algorithm~\ref{alg:gradient2}. It can be shown that both algorithms have a complexity of $O(n_1n_2r)$ per iteration, but empirically the gradient descent with the orthographic retraction is faster since it does not need to compute the singular value decomposition of $\bL^{(k)}$ in each iteration.

\begin{algorithm}[H]
\caption{Gradient descent on the manifold with the projective retraction $R_{\bX}^{(1)}$.}\label{alg:gradient}
\vspace{0.3cm}
{\bf Input:}  Observation $\bY\in\reals^{n_1\times n_2}$; Rank $r$; Thresholding value $\gamma$; Step size $\eta$. \\ 
{\bf Initialization:}  Set $k=0$; Initialize $\bL^{(0)}$ using the rank-$r$ approximation to $f(\bY)$.\\ 
{\bf Loop:} Iterate Steps 1--3 until convergence:
\begin{enumerate}
\item Calculate $P_{T_{\bL^{(k)}}\calM} F(\bL^{(k)}-\bY)$ by \eqref{eq:threshold}, \eqref{eq:derivative} and \eqref{eq:projection}. For the partially-observed case, calculate $P_{T_{\bL^{(k)}}\calM} \tilde{F}(\bL^{(k)}-\bY)$ by \eqref{eq:threshold2}.
\item Let $\bL^{(k+1)}$ be the rank $r$ approximation of $\bL^{(k)} - \eta  P_{T_{\bL^{(k)}}\calM} F(\bL^{(k)}-\bY)$ or $\bL^{(k)} - \eta  P_{T_{\bL^{(k)}}\calM} \tilde{F}(\bL^{(k)}-\bY)$.
\item $k\coloneqq k+1$. 
\end{enumerate}
{\bf Output:} Estimation of the low-rank matrix, given by $\lim_{k\rightarrow\infty}\bL^{(k)}$.
\vspace{0.3cm}
\end{algorithm}

\begin{algorithm}[H]
\caption{\ Gradient descent on the manifold with the orthographic retraction $R_{\bX}^{(2)}$.}\label{alg:gradient2}
\vspace{0.3cm}
{\bf Input:}  Observation $\bY\in\reals^{n_1\times n_2}$; Rank $r$; Thresholding value $\gamma$; Step size $\eta$.\\ 
{\bf Initialization:}  Set $k=0$; Initialize $\bL^{(0)}$ using the rank-$r$ approximation to $f(\bY)$.\\
{\bf Loop:} Iterate Steps 1--3 until convergence:
\begin{enumerate}
\item Calculate $\bD=F(\bL^{(k)}-\bY)$ by \eqref{eq:threshold} for the partially-observed case and $\bD=\tilde{F}(\bL^{(k)}-\bY)$ by \eqref{eq:threshold2} for the partially-observed case. 
\item Let $\bQ\in\reals^{n_1\times r}$ consists of any $r$ independent columns of $\bL^{(k)}$, and $\bR\in\reals^{n_2\times r}$ consists of any $r$ independent rows of $\bL^{(k)}$, and
\[
\bL^{(k+1)}\coloneqq \left[(\bL^{(k)}-\eta \bD)\bR\right]\left[\bQ^{T}(\bL^{(k)}-\eta \bD)\bR\right]^{-1}\left[\bQ^{T}(\bL^{(k)}-\eta \bD)\right].
\]
\item $k\coloneqq k+1$. 
\end{enumerate}
{\bf Output:} Estimation of the low-rank matrix, given by $\lim_{k\rightarrow\infty}\bL^{(k)}$.
\vspace{0.3cm}
\end{algorithm}



\subsection{Partial Observations}\label{sec:partial}
The proposed algorithm can be generalized to the setting of  partial observations, i.e., in addition to gross corruptions,
the data matrix has a large number of missing values. We assume that each entry of $\bY$ is observed with probability $p\in(0,1)$, and denote the set of all observed entries by $\mathbf{\Phi}=\{(i,j)|\bY_{ij}\,\,\text{is observed}\}$. 

For this case, our optimization problem is given by
\[
\argmin_{\rank(\bL)=r}\frac{1}{2}\tilde{f}(\bL),\,\,\,\tilde{f}(\bL)=\sum_{(i,j)\in\mathbf{\Phi}}\tilde{F}_{ij}(\bL-\bY)^2,
\]
where $\tilde{F}$ is defined by
\begin{equation}\label{eq:threshold2}
\tilde{F}_{ij}(\bL-\bY)=\begin{cases}\bL-\bY,\,\,\,\text{if $|[\bL-\bY]_{ij}|>|[\bL-\bY]_{i,\cdot}^{[\gamma,\mathbf{\Phi}]}|$, $|[\bL-\bY]_{ij}|>|[\bL-\bY]_{\cdot,j}^{[\gamma,\mathbf{\Phi}]}|$}\\0,\,\,\,\,\text{otherwise}.\end{cases}
\end{equation}
Here $[\bL-\bY]_{i,\cdot}^{[\gamma,\mathbf{\Phi}]}$ and $|[\bL-\bY]_{ij}|>|[\bL-\bY]_{\cdot,j}^{[\gamma,\mathbf{\Phi}]}|$ represent the $\gamma$-th percentile of the absolute values of the observed entries of the $i$-th row and the $j$-th column respectively.

By a similar argument as in the previous section, we can conclude that when all elements of $|\bL-\bY|$ are different from each other, then applying the same procedure of deriving \eqref{eq:derivative}, we have
\[
\grad\tilde{f}(\bL)=\tilde{F}(\bL-\bY).
\]
Based on the gradient, the algorithm in the partially-observed setting is identical to \eqref{eq:ortho_retraction1}, with $F$ replaced by $\tilde{F}$.

It can be shown that for the partially-observed setting, then the computational cost of Algorithm~\ref{alg:gradient} and Algorithm~\ref{alg:gradient2} is in the order of $O(r^2(n_1+n_2)+r|\mathbf{\Phi}|)$ and the storage is in the order of $O(|\mathbf{\Phi}|+r(n_1+n_2))$.

\section{Theoretical analysis}\label{sec:theory}
In this section, we analyze the theoretical properties of our gradient descent algorithms on the manifold and compare them with previous algorithms. To avoid identifiability issues,  we need to make sure that $\bL^{*}$ can not be both low-rank and sparse. Specifically, we make the following standard assumptions on $\bL^{*}$ and $\bS^{*}$:
\begin{enumerate}
\item  Each row of $\bS^*$ contains at most $\gamma^*n_2$ nonzero entries and each column of $\bS^*$ contains at most $\gamma^*n_1$ nonzero entries. In other words, for $\gamma^{*}\in[0,1)$, assume $\bS^*\in \mathcal{S}_{\gamma^{*}}$ where
\begin{equation}\label{eq:ass1}
\mathcal{S_{\gamma^{*}}\coloneqq}\left\{ A\in\mathbb{R}^{n_{1}\times n_{2}}\mid\|A_{i,\cdot}\|_{0}\leq\gamma^{*} n_{1},\ \mathrm{for}\ 1\leq i\leq n_{1};\|A_{\cdot,j}\|_{0}\leq\gamma^{*} n_{2},\ \mathrm{for}\ 1\leq j\leq n_{2}\right\}. 
\end{equation}
\item The low-rank matrix $\bL^{*}$ is not near-sparse. To achieve this, we require that  $\bL^{*}$ must be $\mu$-coherent. Given the singular value decomposition (SVD)  $\bL^{*}=\bU^{*}\bSigma^{*}\bV^{*T}$, where $\bU^{*}\in\reals^{n_1\times r}$ and  $\bV^{*}\in\reals^{n_2\times r}$, we assume
\begin{equation}\label{eq:ass2}
\|\bU^{*}\|_{2,\infty}\leq \sqrt{\frac{\mu r}{n_1}}, \,\,\,\|\bV^{*}\|_{2,\infty}\leq \sqrt{\frac{\mu r}{n_2}},
\end{equation}
where the norm $\|\cdot\|_{2,\infty}$ is defined by  $\|\bA\|_{2,\infty}=\max_{\|\bz\|_2=1}\|\bA\bz\|_\infty$ and $\|\bx\|_\infty=\max_{i}|\bx_i|$.
\end{enumerate}

With assumptions \eqref{eq:ass1} and \eqref{eq:ass2}, we have the following theoretical results regarding the convergence rate, initialization and stability of Algorithm~\ref{alg:gradient} and Algorithm~\ref{alg:gradient2}:
\begin{thm}[Convergence rate, partially-observed case]\label{thm:main}
The gradient descent algorithms have a linear convergence rate. Suppose that $\|\bL^{(0)}-\bL^*\|_F\leq a\sigma_r(\bL^*)$, where $\sigma_r(\bL^*)$ is the $r$-th largest singular value of $\bL^*$, $a\leq 1/2$, $\gamma>2\gamma^*$ and  $C_1=\sqrt{4(\gamma+2\gamma^*)\mu r+ 4\frac{\gamma^*}{\gamma-\gamma^*} +a^2}<\frac{1}{2}$, then there exists $\eta_0>0$ such that for all $\eta<\eta_0$,
\[
\|\bL^{(k)}-\bL^*\|_F\leq \Big(1-\frac{1-2C_1}{8}\eta\Big)^k \|\bL^{(0)}-\bL^*\|_F.
\]
\end{thm}

\begin{rem}
In particular, if there exists $c_1$ and $c_2$ such that if $a<c_1$, $\gamma^*\mu r<c_2$ and $\gamma=65\gamma^*$, then one can choose $\eta_0=1/8$.
\end{rem}

\begin{thm}[Convergence rate, partially-observed case]\label{thm:main2}
There exists $c>0$ such that for $n=\max(n_1,n_2)$, if $p\geq \max(c\mu r\log(n)/\epsilon^2\min(n_1,n_2),\frac{56}{3}\frac{\log n}{\gamma \min(n_1,n_2)})$, then with probability $1-2n^{-3}-6n^{-1}$, for all $\eta<\eta_0$,
\begin{align*}
\frac{\|\bL^{(k)}-\bL^*\|_F}{\|\bL^{(0)}-\bL^*\|_F}\leq&\Big[\sqrt{1-p^2(1-\epsilon)^2\left(2\eta\left(1-\tilde{C}_1-\frac{ap(1+\epsilon)}{2(1-a)}(1+\tilde{C}_1)\right)-\eta^2(1+\tilde{C}_1)^2\right)}\\+&\frac{\eta^2a^2(p+p\epsilon)^2(1+\tilde{C}_1)^2}{1-\eta a(p+p\epsilon)(1+\tilde{C}_1)}\Big]^k 
\end{align*}
for 
\[
\tilde{C}_1=\frac{1}{p(1-\epsilon)}\Big[6(\gamma+2\gamma^*)p\mu r+4\frac{3\gamma^*}{\gamma-3\gamma^*}(\sqrt{p(1+\epsilon)}+\frac{a}{2})^2+a^2\Big].
\]
\end{thm}
\begin{rem}
In particular, if there exists $\{c_i\}_{i=1}^4>0$ such that when $\epsilon<1/2$, $a<c_1p$, $\gamma^*\mu r<c_2$ and $\gamma=c_3\gamma^*$, then we can choose $\eta_0=1/8$, thus when $\eta<\eta_0$, 
\[
\frac{\|\bL^{(k)}-\bL^*\|_F}{\|\bL^{(0)}-\bL^*\|_F}\leq (1-c_4 \eta p^2)^k.
\]
\end{rem}

Since both statements require proper initializations, the question arises as to how to choose  proper initializations. The work by \cite{DBLP:conf/nips/YiPCC16} shows that if the rank-$r$ approximation to $F(\bY)$ is used as the initialization $\bL^{(0)}$, then such initialization has the upper bound $\|\bL^{(0)}-\bL^*\|$ according to the proofs of \cite[Theorems 1 and 3]{DBLP:conf/nips/YiPCC16} (we borrow this estimation along with the fact that $\|\bL^{(0)}-\bL^*\|_F\leq \sqrt{2r}\|\bL^{(0)}-\bL^*\|$).
\begin{thm}[Initialization, partially-observed case]\label{thm:initialization}
If $\gamma>\gamma^*$ and we initialize $\bL^{(0)}$ using the rank-$r$ approximation to $F(\bY)$, then
\[
\|\bL^{(0)}-\bL^*\|_F\leq 8\gamma \mu r \sqrt{2r}\sigma_1(\bL^*).
\]
\end{thm}
\begin{thm}[Initialization, partially-observed case]\label{thm:initialization2}
There exists $\{c_i\}_{i=1}^3>0$ and $c'>0$ such that if  $\gamma>2\gamma^*$, and $p\geq c_2(\frac{\mu r^2}{\epsilon^2}+\frac{1}{\alpha})\log n/\min(n_1,n_2)$, and we initialize $\bL^{(0)}$ using the rank-$r$ approximation to $f(\bY)$, then
\[
\|\bL^{(0)}-\bL^*\|_F\leq 16 \gamma \mu r \sigma_1(\bL^*)\sqrt{2r} + 2\sqrt{2} c'\epsilon \sigma_1(\bL^*)
\]
with probability at least $1-c_3n^{-1}$, where $\sigma_1(\bL^*)$ is the largest singular value of $\bL^*$
\end{thm}

The combination of Theorem~\ref{thm:main} and \ref{thm:initialization}  implies that, for the partially-observed setting,  the tolerance of the proposed algorithms to corruption is at most  $\gamma^* =O(\frac{1}{\mu r \sqrt{r} \kappa})$, where $\kappa=\sigma_1(\bL^*)/\sigma_r(\bL^*)$ is the conditional number of $\bL^*$. The combination of Theorem~\ref{thm:main2} and \ref{thm:initialization2} implies that, for the partially-observed setting, the proposed algorithm allows the corruption level $\gamma^* =O(\frac{p}{\mu r \sqrt{r} \kappa})$.
 
We also study the stability of the algorithm for the partially-observed case. 
\begin{thm}[Stability, partially-observed case]\label{thm:noisy}
Let $\bL$ be the current value, and let $\bL^+$ be the next update by applying Algorithm~\ref{alg:gradient} or~\ref{alg:gradient2} to $\bL$ for one iteration. Assuming  $\bY=\bL^*+\bS^*+\bN^*$, where $\bN^*$ is a random Gaussian noise i.i.d. sampled from $N(0,\sigma^2)$, $\gamma>10\gamma^*$ and  $(\gamma+2\gamma^*)\mu r < 1/64$, then there exists  $C,a>0$ such that if 
\[
C\sigma \sqrt{(n_1+n_2)r\ln(n_1n_2)}\leq \|\bL-\bL^*\|_F\leq a\sigma_r(\bL^*),
\] then for sufficient small step size $\eta$, we have that
\[
P(\|\bL^+-\bL^*\|_F\leq \left(1-c \eta \right) \|\bL-\bL^*\|_F) \overset{p}{\rightarrow} 1,\quad as\ n_1,n_2\rightarrow\infty,
\]
where $c>0$, thus $0<(1-c\eta)<1$.
\end{thm}
Theorem \ref{thm:noisy} shows that when the observation $\bY$ is contaminated with a random Gaussian noise, if  $\bL^{(0)}$ is properly initialized such that $\|\bL^{(0)}-\bL^*\|_F<a\sigma_r(\bL^*)$, Algorithms~\ref{alg:gradient} and~\ref{alg:gradient2} can converge to a neighborhood of $\bL^*$ given by \[\{\bL:  \|\bL-\bL^*\|_F\leq C\sigma \sqrt{(n_1+n_2)r\ln(n_1n_2)}\}\] with high probability.




\subsection{Comparison with previous works}\label{sec:compare}
Theorems~\ref{thm:main} and \ref{thm:main2} are in parallel with the analysis in \cite{DBLP:conf/nips/YiPCC16}, which is natural since the objective function~\eqref{eq:objective} is equivalent to the one in \cite{DBLP:conf/nips/YiPCC16}. Our methods use the gradient descent on the manifold of low-rank matrices, while the methods in \cite{DBLP:conf/nips/YiPCC16} use the Burer-Monteiro decomposition. In the following we compare the results of both works from four aspects: 
\begin{enumerate}

\item \textbf{Accuracy of initialization.} What is the largest value $t$ that the algorithm can tolerate, such that for any initialization $\bL^{(0)}$ satisfying $\|\bL^{(0)}-\bL^*\|_F\leq t$, the algorithm is guaranteed to converge to $\bL^*$? 

\item \textbf{Convergence rate.} What is the smallest number of iteration steps $k$ such that the algorithm reaches a given convergence criterion $\epsilon$,  i.e. $\|\bL^{(k)}-\bL^*\|_F/\|\bL^{(0)}-\bL^*\|_F<\epsilon$? 

\item \textbf{Corruption level (perfect initialization).} Suppose that the initialization is in a sufficiently small neighborhood of $\bL^*$ (i.e. there exists a very small $\epsilon_0>0$ such that $\bL^{(0)}$ satisfies $\|\bL^{(0)}-\bL^*\|_F<\epsilon_0$), what is the maximum corruption level that can be tolerated in the convergence analysis?  

\item \textbf{Corruption level (proper initialization).} Suppose that the initialization is given by the procedure in Theorem~\ref{thm:initialization} (for the partially-observed case) and \ref{thm:initialization2} (for the partially-observed case), what is the  maximum corruption level that can be tolerated?
\end{enumerate}

These  comparisons are summarized in Table~\ref{tab:compare}. We can see that in the full observed setting, our results remove or reduce the dependence on the conditional number $\kappa$, while keeping other values unchanged. In the partially-observed setting our results still have the advantage of less dependence on $\kappa$, but sometimes require an additional dependence on $p$. The simulation results discussed in the next section also verify that when $\kappa$ is large our algorithms have better performance, while that the slowing effect of $p$ in the partially-observed setting is not significant. 

The results in \cite{NIPS2014_5430} and \cite{DBLP:journals/corr/CherapanamjeriG16} are less comparable since these algorithms are based on the alternative projection instead of the gradient descent. In fact, in the partially-observed case, \cite{NIPS2014_5430} achieves exact recovery when $\gamma=O(1/\mu^2r)$. Compared with $O(1/\mu r^{1.5} \kappa)$  obtained from  combining Theorem~\ref{thm:main} and~\ref{thm:initialization},  \cite{NIPS2014_5430}  removes the dependence on $\kappa$ and reduces the dependence on $r$, but requires a stronger dependence on $\mu$. In addition, the  algorithm in  \cite{NIPS2014_5430}  has a complexity of $O(n_1n_2r^2)$. It is slightly more expensive than $O(n_1n_2r)$, which is the complexity of our algorithms and the algorithms in \cite{DBLP:conf/nips/YiPCC16}.

\begin{table*}[htbp]
\centering
\begin{tabular}{lllll}
\toprule 
Criterion  & 1  & 2  & 3  & 4\tabularnewline
\midrule 
Proposed method (full)  & $O(\sigma_{r}(\bL^{*}))$  & $O(\log(\frac{1}{\epsilon}))$  & $O(\frac{1}{\mu r})$ & $O(\frac{1}{\mu r^{1.5}\kappa})$\tabularnewline
\cite{DBLP:conf/nips/YiPCC16} (full)  & $O(\frac{\sigma_{r}(\bL^{*})}{\sqrt{\kappa}})$  & $O(\kappa\log(\frac{1}{\epsilon}))$  & $O(\frac{1}{\kappa^{2}\mu r})$ & $O(\frac{1}{\max(\mu r^{1.5}\kappa^{1.5},\kappa^{2}\mu r)})$\tabularnewline
Proposed method (partial)  & $O(p\sigma_{r}(\bL^{*}))$  & $O(\log(\frac{1}{\epsilon})/p^{2})$  & $O(\frac{1}{\mu r})$  & $O(\frac{p}{\mu r^{1.5}\kappa})$\tabularnewline
\cite{DBLP:conf/nips/YiPCC16} (partial)  & $O(\frac{\sigma_{r}(\bL^{*})}{\kappa})$  & $O(\kappa\mu r\log(\frac{1}{\epsilon})$  & $O(\frac{1}{\kappa^{2}\mu r})$  & $O(\frac{1}{\max(\mu r^{1.5}\kappa^{1.5},\kappa^{2}\mu r)})$\tabularnewline
\bottomrule
\end{tabular}

\caption{\small {Comparison of the theoretical guarantees in our work and in \cite{DBLP:conf/nips/YiPCC16}. The four criteria are explained in details in Section~\ref{sec:compare}.  \label{tab:compare}}}

\end{table*}


\begin{figure}
\centering
\includegraphics[width=0.49\textwidth]{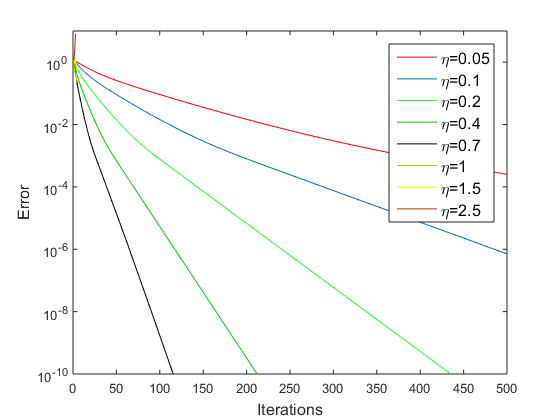}
\includegraphics[width=0.49\textwidth]{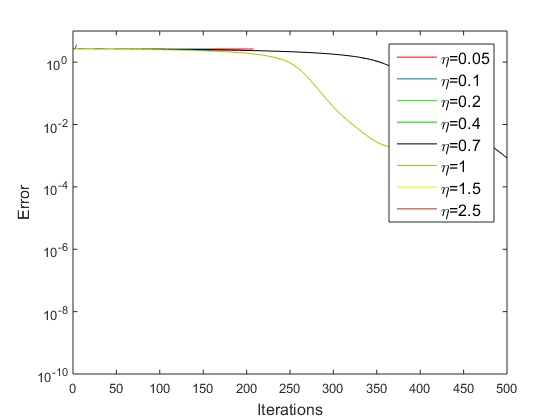}

\centering
\includegraphics[width=0.49\textwidth]{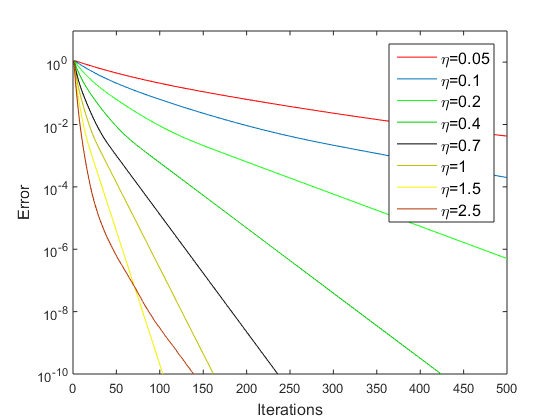}
\includegraphics[width=0.49\textwidth]{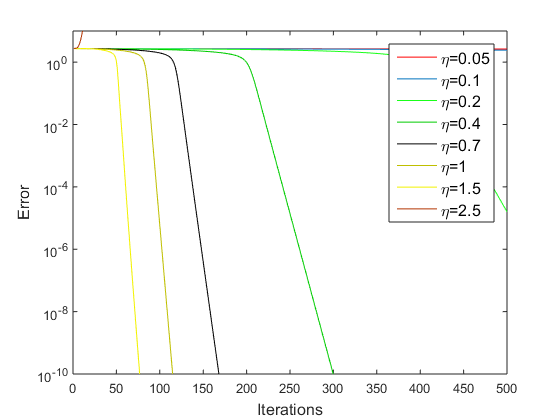}

\centering
\includegraphics[width=0.49\textwidth]{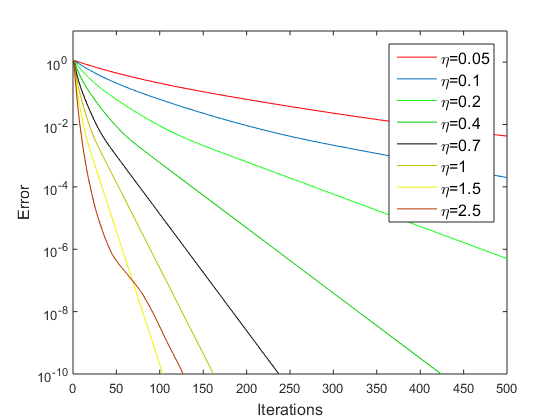}
\includegraphics[width=0.49\textwidth]{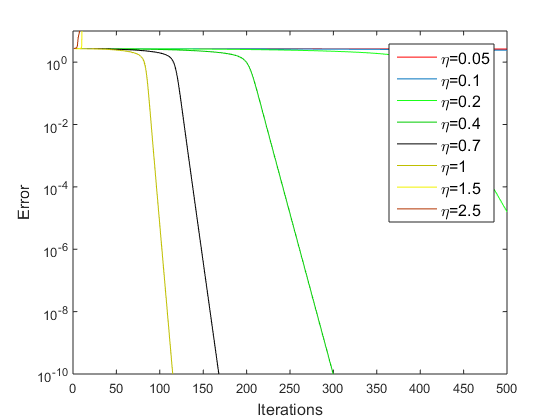}
\caption{The dependence of the estimation error with respect to the number of iterations for Algorithm~\ref{alg:gradient} (partially-observed case) with various choices of step sizes. Left column: Setting 1. Right column: Setting 2. First row: the method in \cite{DBLP:conf/nips/YiPCC16}. Second row: Algorithm~\ref{alg:gradient}. Third row: Algorithm~\ref{alg:gradient2}. The step sizes are chosen among $\eta \in \{0.05,0.1,0.2,0.4,0.7,1,1.5,2.5\}$.}\label{fig:DD1}
\end{figure}

\begin{figure}
\centering
\includegraphics[width=0.49\textwidth]{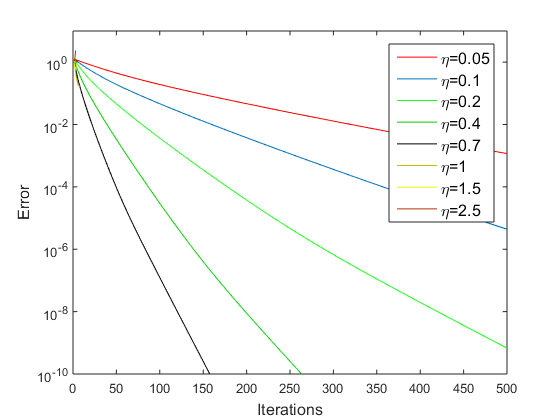}
\includegraphics[width=0.49\textwidth]{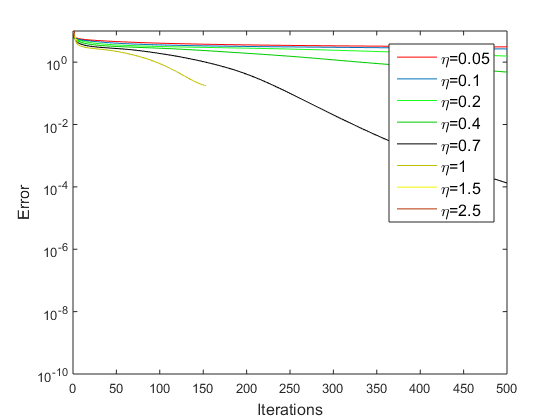}
\centering
\includegraphics[width=0.49\textwidth]{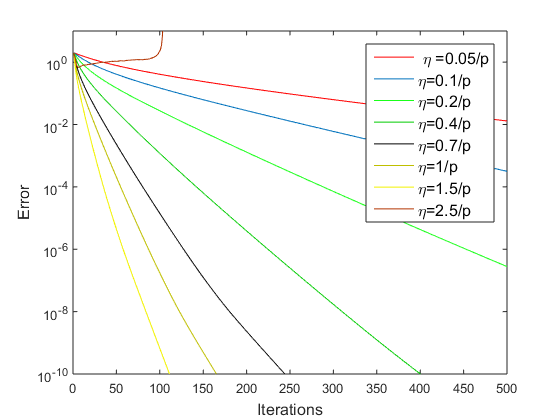}
\includegraphics[width=0.49\textwidth]{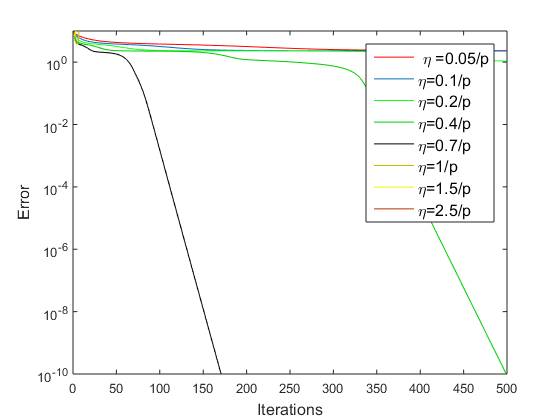}
\centering
\includegraphics[width=0.49\textwidth]{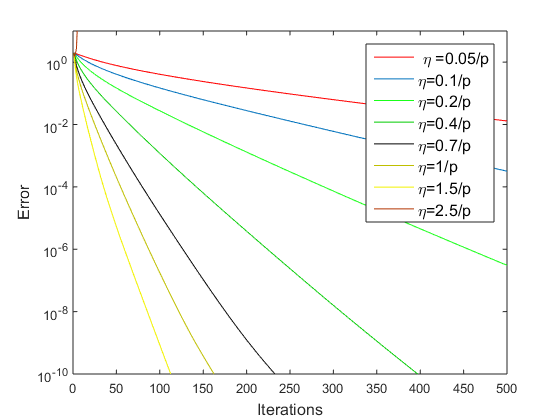}
\includegraphics[width=0.49\textwidth]{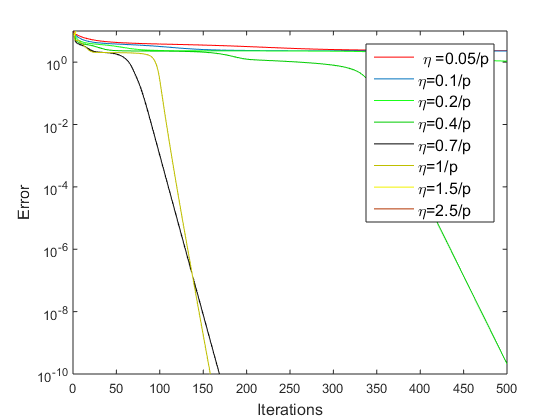}

\caption{The dependence of the estimation error with respect to the number of iterations for Algorithm~\ref{alg:gradient2} (partially-observed case) with various choices of step sizes. Left column: Setting 1, partially-observed case with $p=0.2$. Right column: partially-observed case with $p=0.2$. First row: the method in \cite{DBLP:conf/nips/YiPCC16}. Second row: Algorithm~\ref{alg:gradient}. Third row: Algorithm~\ref{alg:gradient2}. The step sizes are chosen among $\eta \in \{0.05,0.1,0.2,0.4,0.7,1,1.5,2.5\}$.}\label{fig:DD2}
\end{figure}
 
\section{Simulations}\label{sec:simu}
In this section, we compare the computational performances of our method and the method in \cite{DBLP:conf/nips/YiPCC16}. In  simulations, we let $[n_1,n_2]=[500,600]$, $r=5$, and $\bL^*$ is generated from $\bU\bSigma\bV^T$, where $\bU$ and $\bV$ are random orthogonal matrices of size $\reals^{500\times 5}$ and $\reals^{600\times 5}$. We consider the following two settings:
\begin{itemize}
\item \textbf{Setting 1.} $\bSigma=\bI$ (the condition number $\kappa$ is thus 1), $\bY$ is obtained by replacing $25$ elements in each column of $\bL^*$ by a random number from Gaussian distribution $\mathcal{N}(0,1)$, and let $\gamma=0.2$. 
\item \textbf{Setting 2.}  $\bSigma=\diag([10,1,1,1,1])$ (the condition number $\kappa$ is thus 10), $\bY=\bL^{*}$, and let $\gamma=0.05$.
\end{itemize}

The performance of the algorithms with various choices of step sizes are recorded in Figure~\ref{fig:DD1} and \ref{fig:DD2}, where the error is measured by the Frobenius norm of the difference to the underlying low-rank matrix, i.e. $\|\bL^{(i)}-\bL^*\|_F$ for all $i$. For all cases of the simulation results, the algorithms usually converge faster with larger step sizes, but will diverge once reach a certain threshold. Therefore in all simulations we test a wide range of step sizes $\eta \in \{0.05,0.1,0.2,0.4,0.7,1,1.5,2.5\}$, so that the best step sizes are included. For example, for the upper-left figure in Figure~\ref{fig:DD1} and \ref{fig:DD2}, the algorithm converges when $\eta$ is between $0.05$ and $0.7$, and diverges when $\eta\geq 1$, and $\eta=0.7$ is the best step size in terms of convergence rate. For the partially-observed setting, we use the step sizes with a factor $1/p$ as it works well empirically. 

Figure~\ref{fig:DD1} and \ref{fig:DD2} show that our algorithms converge linearly, and faster than the algorithm in \cite{DBLP:conf/nips/YiPCC16} when the step sizes are well-chosen.  The performance our algorithms is also less sensitive to the choice of step sizes. For both Setting 1 and 2, both partially-observed cases and partially-observed cases, our algorithms converge within $300$ iterations for a wide range of $\lambda$.  The advantage of our algorithms are more obvious when the conditional number is large. In particular, the right columns of Figure~\ref{fig:DD1} and \ref{fig:DD2} visualize the cases for Setting 2 when the conditional number of $\bL^*$ is $10$. Under these cases, the advantage of our algorithms in convergence rate becomes more obvious. This verifies the analysis in Section~\ref{sec:compare} that our algorithms remove or reduce the dependence of the convergence on the conditional number $\kappa$. In addition, we observed that the performance of our algorithms in the partially-observed case is not significantly affected by the presence of the additional dependence on the observation probability $p$.

In addition, the simulations shows that the projective retraction and the orthographic retraction usually has no significant difference in terms of performance. In fact, their performance are almost identical except in Figure \ref{fig:DD1}, when $\eta=1.5$ for partially-observed case and in Figure \ref{fig:DD2}, when $\eta=1$ for partially-observed case, both under Setting 2. Since it is less computationally extensive to calculate the orthographic retraction (the projective retraction requires an additional singular value decomposition in each iteration), we recommend to use Algorithm~\ref{alg:gradient2}, especially when the computational complexity is concerned.

\begin{figure}
\centering\includegraphics[width=1.0\textwidth]{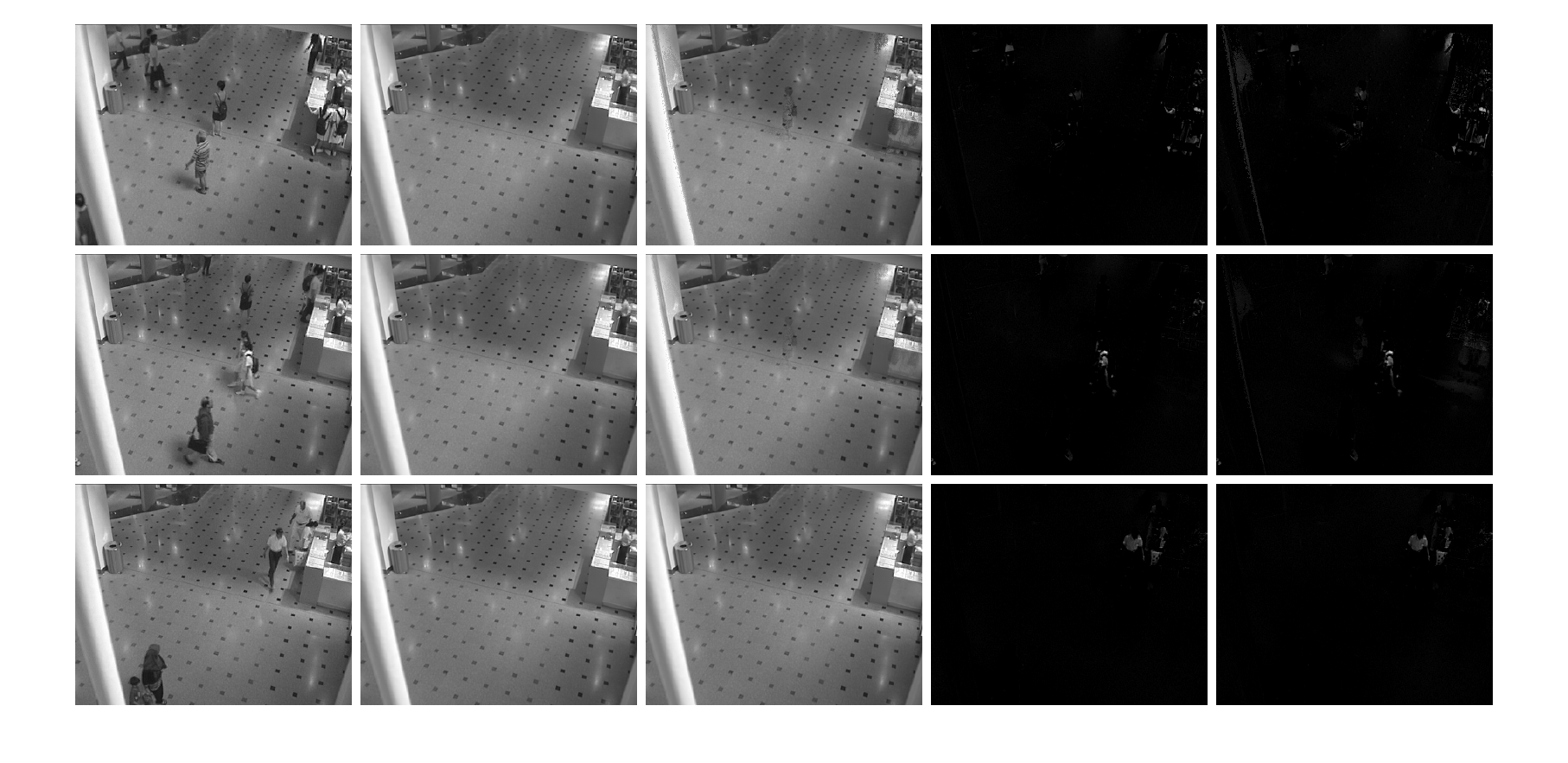}
\caption{The performance of Algorithm~\ref{alg:gradient2} in video background subtraction, with three rows representing three frames in the video sequence. Five columns represent (from left to right) the original frame, the background by Algorithm~\ref{alg:gradient2} in partially-observed case, the background by Algorithm~\ref{alg:gradient2} in partially-observed case with $p=0.5$, the foreground by Algorithm~\ref{alg:gradient2} in partially-observed case, and the foreground  by Algorithm~\ref{alg:gradient2} in partially-observed case.}\label{fig:DD3}
\end{figure}

\begin{figure}
\centering\includegraphics[width=0.49\textwidth]{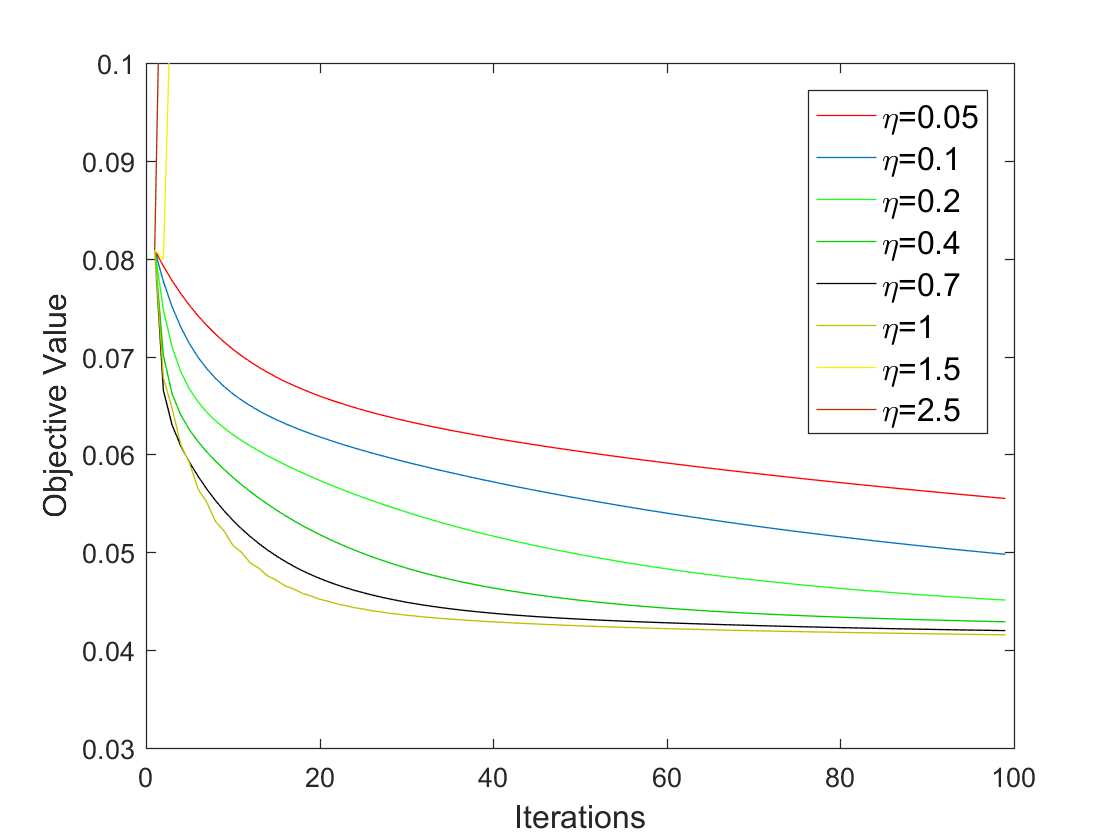}\includegraphics[width=0.49\textwidth]{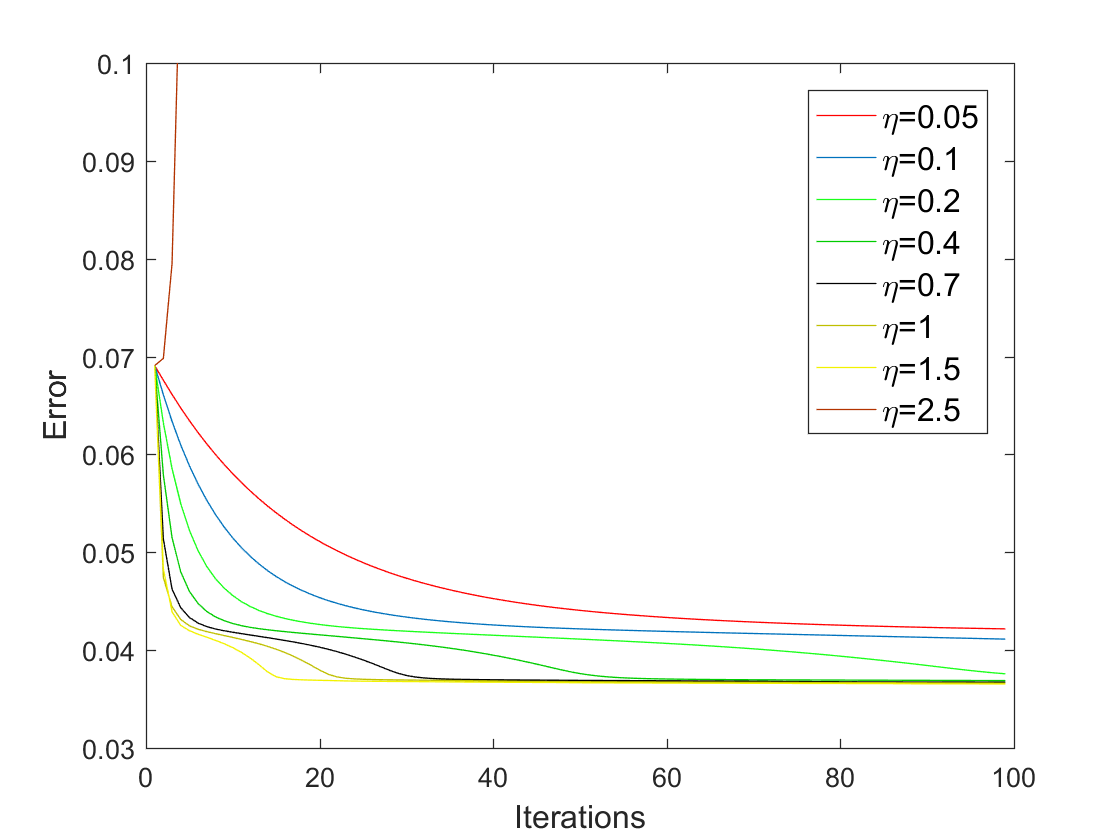}

\centering\includegraphics[width=0.49\textwidth]{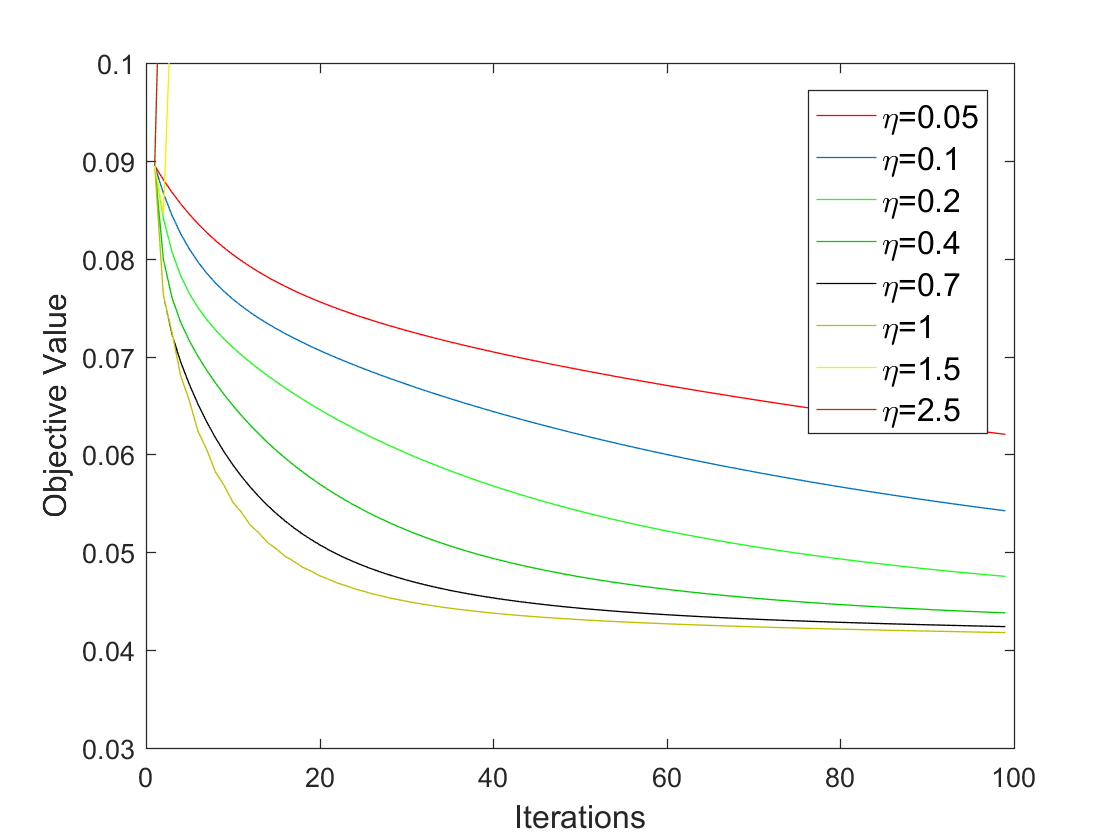}\includegraphics[width=0.49\textwidth]{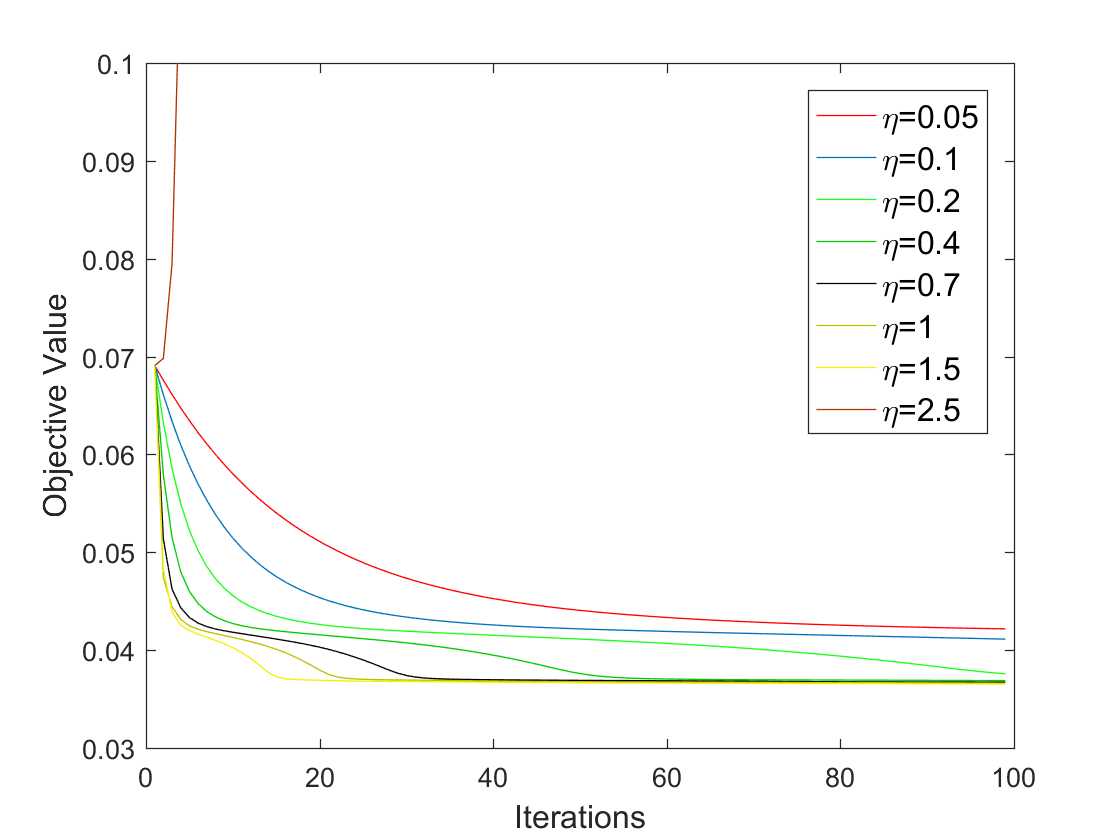}

\caption{The value of the objective function with respect to the iterations for various choices of step sizes. Top row: partially-observed case. Second row: partially-observed case.  Left: the method in \cite{DBLP:conf/nips/YiPCC16}.  Right: Algorithm~\ref{alg:gradient2}.}\label{fig:DD4}
\end{figure}

We test Algorithm~\ref{alg:gradient2} in a real data application for video background subtraction. We adopt the public data set \texttt{Shoppingmall} studied in \cite{DBLP:conf/nips/YiPCC16},\footnote{The data set is originally from \url{http://perception.i2r.a-star.edu.sg/bk_model/bk_index.html}, and is available at \url{https://sciences.ucf.edu/math/tengz/}.} A few frames are visualized in the first column of Figure~\ref{fig:DD3}. There are 1000 frames in this video sequence, represented by a matrix of size $81920\times 1000$, where each column corresponds to a frame of the video and each row corresponds to a pixel of the video. We apply Algorithm~\ref{alg:gradient2} with $r=3$ and $\gamma^*=0.1$, $p=0.5$ for the partially-observed case, the step size $\eta=0.7$. We stop the algorithm after 100 iterations. Figure~\ref{fig:DD3} shows that our algorithms obtain desirable low-rank approximations within $100$ iterations. In Figure~\ref{fig:DD4}, we also compare Algorithm~\ref{alg:gradient2} and the method in \cite{DBLP:conf/nips/YiPCC16} with respect to the convergence of the objective function value, and we expect that a smaller objective function implies a better low-rank approximation. It turns out that our algorithm can consistently obtain smaller objective value within 100 iterations under both fully-observed and partially-observed settings.


\section{Conclusion}\label{sec:discussion}
In this paper we propose two robust PCA algorithms (one for projective retraction and one for orthographic retraction) based on the gradient descent algorithm on the manifold of low-rank matrices. Theoretically, compared with the gradient descent algorithm with Burer-Monteiro decomposition, our approach has a faster convergence rate, better tolerance of the initialization accuracy and corruption level. The approach removes or reduces the dependence of the algorithms on the conditional number of the underlying low-rank matrix.  Numerically, the proposed algorithms performance is less sensitive to the choice of step sizes. We also find that under the partially-observed setting, the performance of the proposed algorithm is not significantly affected by the presence of the additional dependence on the observation probability.  Considering the popularity of the Burer-Monteiro decomposition, it is an interesting future direction to apply manifold optimization to other low-rank matrix estimation problems.

\newpage
\section*{Appendix for ``Robust Principal Component Analysis by Manifold Optimization"}
\subsection*{A. Proof of Theorem~\ref{thm:main}}
In this proof, we will investigate $\|\bL^+-\bL^*\|_F$, where 
\[
\bL^+=R_{\bL}(-\eta P_{T_{\bL}} F(\bL-\bY)).
\]
It is sufficient to prove that when $\|\bL-\bL^*\|\leq a \sigma_r(\bL^*)$ with $a$ satisfying the conditions in Theorem~\ref{thm:main}, then
\begin{equation}\label{eq:goal1}
\|\bL^+-\bL^*\|_F\leq \Big(1-\frac{1-2C_1}{8}\eta\Big) \|\bL-\bL^*\|_F.
\end{equation}
To prove \eqref{eq:goal1}, we first introduce three auxiliary lemmas.
\begin{lem}\label{lemma:bD}
\begin{description}
\item (a) Let $\bD=\bL-\bL^*-F(\bL-\bY)=\bL-\bL^*-F(\bL-\bL^*-\bS^*)$, then
\begin{equation}\label{eq:second_bound}
\|\bD\|_F^2\leq C_1^2\|\bL-\bL^*\|_F^2.
\end{equation}

\item (b) For the noisy setting where $\bY=\bL^*+\bS^*+\bN^*$, and $\bD'=\bL-\bL^*-\bN^*-F(\bL-\bY)$, we have
\begin{equation}\label{eq:second_bound_noisy}
\|\bD'\|_F^2\leq 2C_1^2\|\bL-\bL^*\|_F^2+2(\gamma+5\gamma^*)N_1,
\end{equation}
where $N_1=n_2\sum_{i=1}^{n_1}[\bN^*_{i,\cdot}]^{(1)}+n_1\sum_{j=1}^{n_2}[\bN^*_{\cdot,j}]^{(1)}$. 
\end{description}
\end{lem}

\begin{lem}\label{lemma:approximation}
If $\|\bL-\bL^*\|_F= a\sigma_r(\bL^*)$ and $a\leq 1$, then 
\begin{align}
\|(\bL-\bL^*)-\bP_{\mathcal{T}_{\bL}}(\bL-\bL^*)\|_F\leq &\frac{a}{2(1-a)}\|\bL-\bL^*\|_F,\label{eq:control1}\\
\|(\bL-\bL^*)-\bP_{\mathcal{T}_{\bL^*}}(\bL-\bL^*)\|_F\leq &\frac{a}{2}\|\bL-\bL^*\|_F\label{eq:control2}.
\end{align}
\end{lem}
\begin{lem}\label{lemma:retraction}
For $\bX\in T_{\bL}\calM$, then
\[
\|R^{(i)}_{\bL}(\bX)-(\bL+\bX)\|_F\leq \frac{\|\bX\|_F^2}{2(\sigma_r(\bL)-\|\bX\|)},\,\,\text{for both $i=1$ or $2$.}
\]
\end{lem}
To prove \eqref{eq:goal1}, first we note that
\begin{align}\nonumber
&\|\bL-\bL^*\|_F^2-\|\bL-\eta P_{T_{\bL}} F(\bL-\bY)-\bL^*\|_F^2 \\\nonumber
=&2 \eta\langle\bL-\bL^*,P_{T_{\bL}} F(\bL-\bY)\rangle-\|\eta P_{T_{\bL}} F(\bL-\bY)\|_F^2.\\\nonumber
=&2 \eta\langle P_{T_{\bL}}(\bL-\bL^*),P_{T_{\bL}}(\bL-\bL^*)-P_{T_{\bL}}\bD\rangle-\eta^2\|P_{T_{\bL}} F(\bL-\bY)\|_F^2\\
=&2\eta (\|P_{T_{\bL}}(\bL-\bL^*)\|_F^2-\|\bD\|_F\|P_{T_{\bL}}(\bL-\bL^*)\|_F)-\eta^2(\|\bL-\bL^*\|_F+\|\bD\|_F)^2.
\label{eq:condition1}
\end{align}
Lemma~\ref{lemma:approximation} and the assumptions $\|\bL-\bL^*\|_F\leq a\sigma_r(\bL^*)$, $\sqrt{1-(\frac{a}{2(1-a)})^2}>\frac{1}{2}$ imply
\begin{equation}
\|P_{T_{\bL}} (\bL-\bL^*)\|_F\geq \frac{1}{2}\|\bL-\bL^*\|_F\label{eq:condition1.5}.
\end{equation}
Combining it with the estimation of $\|\bD\|_F$ in Lemma~\ref{lemma:bD}, we have
\begin{align}
\|\bL-\bL^{*}\|_{F}^{2}-\|\bL-\eta P_{T_{\bL}}F(\bL-\bY)-\bL^{*}\|_{F}^{2}\nonumber\\
\geq\eta(\frac{1}{2}-C_{1})\|\bL-\bL^{*}\|_{F}^{2}-\eta^{2}(1+C_{1})^{2}\|\bL-\bL^{*}\|_{F}^{2}.\label{eq:condition1.1}
\end{align}
When ths RHS of \eqref{eq:condition1.1} is positive (i.e., when $(1-2C_1)\geq 2(1+C_1)^2\eta$), \eqref{eq:condition1.1} implies $\|\bL-\bL^*\|_F>\|\bL-\eta P_{T_{\bL}} F(\bL-\bY)-\bL^*\|_F$ and 
\begin{align}\nonumber
 & \|\bL-\bL^{*}\|_{F}-\|\bL-\eta P_{T_{\bL}}F(\bL-\bY)-\bL^{*}\|_{F}\\
\geq & \frac{\eta(\frac{1}{2}-C_{1})\|\bL-\bL^{*}\|_{F}^{2}-\eta^{2}(1+C_{1})^{2}\|\bL-\bL^{*}\|_{F}^{2}}{\|\bL-\bL^{*}\|_{F}+\|\bL-\eta P_{T_{\bL}}F(\bL-\bY)-\bL^{*}\|_{F}}\nonumber\\
\geq & \frac{1}{2}\left(\eta(\frac{1}{2}-C_{1})-\eta^{2}(1+C_{1})^{2}\right)\|\bL-\bL^{*}\|_{F}.\label{eq:condition1.2}
\end{align}
In addition, \begin{equation}
\|P_{T_{\bL}} F(\bL-\bY)\|_F\leq \|F(\bL-\bY)\|_F= \|\bL-\bL^*\|_F+\|\bD\|_F\leq (1+C_1)\|\bL-\bL^*\|_F\label{eq:condition3.5}\end{equation} and Lemma~\ref{lemma:retraction} give
\begin{align}\nonumber
&\|\bL^+-\bL\|_F-\|\bL-\eta P_{T_{\bL}} F(\bL-\bY)-\bL^*\|_F\leq \|\bL-\eta P_{T_{\bL}} F(\bL-\bY)-\bL^+\|_F\\\leq& \frac{\eta^2\|P_{T_{\bL}} F(\bL-\bY)\|_F^2}{\sigma_r(\bL^*)-\eta\|P_{T_{\bL}} F(\bL-\bY)\|_F}\leq\frac{\eta^2a^2(1+C_1)^2}{1-\eta a(1+C_1)}\|\bL-\bL^*\|_F.\label{eq:condition4}
\end{align}
Combining \eqref{eq:condition1.2} and \eqref{eq:condition4},
\begin{align*}
&\frac{\|\bL-\bL^*\|_F-\|\bL^+-\bL^*\|_F}{\|\bL-\bL^*\|_F}\geq  \frac{1}{4} \eta(1-2C_1)-\eta^2(1+C_1)^2\left[\frac{1}{2}+\frac{a^2}{1-\eta(1+C_1)a}\right].
\end{align*}
Therefore, Theorem~\ref{thm:main} is proved when $C_1<1/2$, and $\eta$ is chosen such that
\[
\eta(1+C_1)^2\left[\frac{1}{2}+\frac{a^2}{1-\eta(1+C_1)a}\right]\leq \frac{1}{8} (1-2C_1).
\]

\subsection*{B. Proof of Theorem~\ref{thm:main2}}
This proof borrows two lemmas from \cite[lemma 9, 10]{DBLP:conf/nips/YiPCC16} as follows.
\begin{lem}\label{lemma:concentration}{\cite[Lemma 9]{DBLP:conf/nips/YiPCC16}} 
There exists $c>0$ such that for all $0<\epsilon<1$, if $p\geq c\mu r\log(n)/\epsilon^2\min(n_1,n_2)$, then with probability at least $1-2n^{-3}$, for any $\bx\in T(\bL^*)$
\[
(1-\epsilon)\|\bX\|_F^2\leq \frac{1}{p}\|P_{\mathbf{\Phi}}\bX\|_F^2\leq  (1+\epsilon)\|\bX\|_F^2.
\]
\end{lem}
\begin{lem}\label{lemma:sampling}{\cite[Lemma 10]{DBLP:conf/nips/YiPCC16}} 
 If $p\geq \frac{56}{3}\frac{\log n}{\gamma \min(n_1,n_2)}$, the with probability at least $1-6n^{-1}$, the number of entries in $\mathbf{\Phi}$ per row is in the interval $[pn_2/2,3pn_2/2]$, and the number of entries in $\mathbf{\Phi}$ per column is in $[pn_1/2,3pn_1/2]$.
\end{lem}

Then we introduce the following lemma parallel to Lemma~\ref{lemma:bD}:

\begin{lem}\label{lemma:bD2}
When the events in Lemma~\ref{lemma:concentration} and~\ref{lemma:sampling} holds, for  $\tilde{\bD}=P_{\mathbf{\Phi}}[\bL-\bL^*-\tilde{F}(\bL-\bY)]$ we have
\begin{equation}\label{eq:second_bound2}
\|\tilde{\bD}\|_F^2\leq \tilde{C}_1^2\|\bL-\bL^*\|_F^2,
\end{equation}
with \[\tilde{C}_1=\frac{1}{p(1-\epsilon)}\Big[6(\gamma+2\gamma^*)p\mu r+4\frac{3\gamma^*}{\gamma-3\gamma^*}(\sqrt{p(1+\epsilon)}+\frac{a}{2})^2+a^2\Big].\]
\end{lem}

The proof of Theorem~\ref{thm:main2} is parallel to the proof of Theorem~\ref{thm:main}, with $\bL^+$ defined slightly differently by
\[
\bL^+=R_{\bL}(-\eta P_{T_{\bL}}\tilde{F}(\bL-\bY)).
\]
Defining $P_{\mathbf{\Phi}}: \reals^{n_1\times n_2}\rightarrow \reals^{n_1\times n_2}$ by
\[
[P_{\mathbf{\Phi}} \bX]_{ij}=\begin{cases}\bX_{ij},\,\,\text{if $(i,j)\in\mathbf{\Phi}$,}\\0,\,\,\text{if $(i,j)\notin\mathbf{\Phi}$.}\end{cases}
\]
Then $\tilde{F}(\bL-\bY)=P_{\mathbf{\Phi}}\tilde{F}(\bL-\bY)$. Following a similar analysis as \eqref{eq:condition1}, 
\begin{align}\nonumber
&\|\bL-\bL^*\|_F^2-\|\bL-\eta P_{T_{\bL}}P_{\mathbf{\Phi}} \tilde{F}(\bL-\bY)-\bL^*\|_F^2 \\\nonumber
=&2 \eta\langle\bL-\bL^*,P_{T_{\bL}}P_{\mathbf{\Phi}} \tilde{F}(\bL-\bY)\rangle-\|\eta P_{T_{\bL}}P_{\mathbf{\Phi}} \tilde{F}(\bL-\bY)\|_F^2\\\nonumber
\geq & 2 \eta\langle P_{\mathbf{\Phi}} P_{T_{\bL}}(\bL-\bL^*),P_{\mathbf{\Phi}} \tilde{F}(\bL-\bY)\rangle-\|\eta P_{\mathbf{\Phi}} \tilde{F}(\bL-\bY)\|_F^2\\
\geq & 2 \eta\langle P_{\mathbf{\Phi}}(\bL-\bL^*)-P_{\mathbf{\Phi}} P_{T_{\bL}}^\perp(\bL-\bL^*),P_{\mathbf{\Phi}}(\bL-\bL^*)-\tilde{\bD}\rangle-\eta^2(\| P_{\mathbf{\Phi}}(\bL-\bL^*)\|_F+\|\tilde{\bD}\|_F)^2
.\label{eq:condition1c}
\end{align}
Lemma~\ref{lemma:approximation} and Lemma~\ref{lemma:concentration} implies
\[
\frac{\|P_{\mathbf{\Phi}} P_{T_{\bL}}^\perp(\bL-\bL^*)\|_F}{\|P_{\mathbf{\Phi}} (\bL-\bL^*)\|_F}\leq \frac{\|P_{T_{\bL}}^\perp(\bL-\bL^*)\|_F}{\|P_{\mathbf{\Phi}} (\bL-\bL^*)\|_F}\leq  \frac{ap(1+\epsilon)}{2(1-a)},
\]
and combining it with the estimation of $\tilde{\bD}$ in Lemma~\ref{lemma:bD2}, the RHS of \eqref{eq:condition1c} is larger than
\begin{equation}
\|P_{\mathbf{\Phi}}(\bL-\bL^*)\|_F^2\left(2\eta\left(1-\tilde{C}_1-\frac{ap(1+\epsilon)}{2(1-a)}(1+\tilde{C}_1)\right)-\eta^2(1+\tilde{C}_1)^2\right).\label{eq:condition1c2}
\end{equation}
In addition,  Lemma~\ref{lemma:concentration} implies
\begin{align*}
\|P_{\mathbf{\Phi}}\tilde{F}(\bL-\bY)\|_{F} & \leq\|P_{\mathbf{\Phi}}(\bL-\bL^{*})\|_{F}+\|P_{\mathbf{\Phi}}\tilde{\bD}\|_{F}\\
 & \leq(1+\tilde{C}_{1})\|P_{\mathbf{\Phi}}(\bL-\bL^{*})\|_{F},\\
 & \leq(1+\tilde{C}_{1})p(1+\epsilon)\|\bL-\bL^{*}\|
\end{align*}
and combining it with Lemma~\ref{lemma:retraction},
\[
\|\bL^+-\bL^*\|_F-\|\bL-\eta P_{T_{\bL}} P_{\mathbf{\Phi}}\tilde{F}(\bL-\bY)-\bL^*\|_F\leq 
\frac{\eta^2a^2(p+p\epsilon)^2(1+\tilde{C}_1)^2}{1-\eta a(p+p\epsilon)(1+\tilde{C}_1)}\|\bL-\bL^*\|_F.
\]
Combining it with \eqref{eq:condition1c2} and Lemma~\ref{lemma:approximation}, we have
\begin{align*}
\frac{\|\bL^+-\bL^*\|_F}{\|\bL-\bL^*\|_F}\leq&\sqrt{1-p^2(1-\epsilon)^2\left(2\eta\left(1-\tilde{C}_1-\frac{ap(1+\epsilon)}{2(1-a)}(1+\tilde{C}_1)\right)-\eta^2(1+\tilde{C}_1)^2\right)}\\+&\frac{\eta^2a^2(p+p\epsilon)^2(1+\tilde{C}_1)^2}{1-\eta a(p+p\epsilon)(1+\tilde{C}_1)},
\end{align*}
and Theorem~\ref{thm:main2} is proved.

\subsection*{C. Proof of Theorem~\ref{thm:noisy}}
The proof of the noisy case also follows similarly from the proofs of Theorem~\ref{thm:main} and~\ref{thm:main2}. Note that
\[
F(\bL-\bY)=\bL-\bL^*-\bN^*-\bD',
\]
and define $\bQ=P_{T_{\bL}}(\bL-\bL^*)$, then following the proof of Theorem~\ref{thm:main} and applying Lemma~\ref{lemma:bD} (b), we have
\begin{align*}\nonumber
&\|\bL-\bL^*\|_F^2-\|\bL-\eta P_{T_{\bL}} F(\bL-\bY)-\bL^*\|_F^2 \\\nonumber
=&2 \eta\langle\bL-\bL^*,P_{T_{\bL}} F(\bL-\bY)\rangle+O(\eta^2)=2 \eta\langle P_{T_{\bL}} (\bL-\bL^*),P_{T_{\bL}} F(\bL-\bY)\rangle+O(\eta^2)\\
=&2 \eta\langle P_{T_{\bL}} (\bL-\bL^*),P_{T_{\bL}} (\bL-\bL^*-\bN^*-\bD')\rangle+O(\eta^2)\\
\geq &2\eta\left(\|\bQ\|_F^2-\langle\bN^*,\bQ\rangle-\|\bQ\|_F\sqrt{2C_1^2\|\bL-\bL^*\|_F^2+2(\gamma+5\gamma^*)N_1}\right)+O(\eta^2).
\end{align*}
In addition, \eqref{eq:condition4} gives \begin{align}\nonumber
\left|\|\bL^+-\bL\|_F-\|\bL-\eta P_{T_{\bL}} F(\bL-\bY)-\bL^*\|_F\right|=O(\eta^2).
\end{align}
Combining it with the estimation of $C_1$, $N_1$, and  $\langle\bN^*,\bQ\rangle$ in Lemma~\ref{lemma:prob} and the fact that $(1-\frac{a}{2(1-a)})\|\bL-\bL^*\|_F\leq \|\bQ\|_F\leq (1+\frac{a}{2(1-a)})\|\bL-\bL^*\|_F$ (which follows from Lemma~\ref{lemma:approximation}), the Theorem is proved. 


\begin{lem}\label{lemma:prob}
If $\bN^*\in\reals^{n_1\times n_2}$ is elementwisely i.i.d. sampled from $N(0,\sigma^2)$, then\\
(a) with probability $1-\frac{4}{n_1^7n_2^7}$, $\sum_{i=1}^{n_1}[\bN^*_{i,\cdot}]^{(1)2}\leq 16\sigma^2 n_1\ln (n_1n_2)$, and $\sum_{j=1}^{n_2}[\bN^*_{\cdot,j}]^{(1)2}\leq 16\sigma^2 n_2\ln (n_1n_2)$, and as a result, $N_1\leq 32\sigma^2n_1n_2\ln (n_1n_2)$.\\
(b) There exists $C_6>0$ such that as $n_1+n_2\rightarrow\infty$, the probability that
\begin{equation}\label{eq:event_noisy}\langle\bN^*, P_{T_{\bL}} (\bL-\bL^*)\rangle\leq \frac{1}{4}\|P_{T_{\bL}} (\bL-\bL^*)\|_F^2\end{equation} holds for all $\{\bL: C_6\sigma\sqrt{(n_1+n_2)r\ln(n_1n_2)}\leq \|\bL-\bL^*\|_F\leq a\sigma_r(\bL^*)\}$ converges to 1.
\end{lem}

\subsection*{D. Proof of Lemmas}
\paragraph*{Proof of Lemma~\ref{lemma:bD}(a)}
\begin{proof}
By the definition of $f$, $\bD$ is a sparse matrix. Denote the locations of the nonzero entries by $\mathcal{S}$, and divide it into two sets $\mathcal{S}_1\cup\mathcal{S}_2$ as follows:
\[
\mathcal{S}_{1}=\{(i,j):|[\bL-\bL^{*}-\bS^{*}]_{ij}|>|[\bL-\bL^{*}-\bS^{*}]_{i,\cdot}^{(\gamma n_{2})}|\ \textrm{and}\ |[\bL-\bL^{*}-\bS^{*}]_{ij}|>|[\bL-\bL^{*}-\bS^{*}]_{\cdot,j}^{(\gamma n_{1})}|\},\] 
and 
\[
\mathcal{S}_2=\{(i,j)\notin\mathcal{S}_1: [\bL-\bL^*]_{ij}\neq F(\bL-\bL^*-\bS^*)_{ij}\}.
\]

For $(i,j)\in\mathcal{S}_1$, $[F(\bL-\bL^*-\bS^*)]_{ij}=0$. As a result, $\bD_{ij}=[\bL-\bL^*]_{ij}$.
In addition, by definition, each row or column has at most $\gamma$ percentage of points in $\mathcal{S}_1$.

For $(i,j)\in\mathcal{S}_2$, $\bS^*_{ij}\neq 0$. Therefore, for each row or column, at most $\gamma^*$ percentage of points lie in $\mathcal{S}_2$. Since $[F(\bL-\bL^*-\bS^*)]_{ij}=[\bL-\bL^*-\bS^*]_{ij}$, and 
\[
|[\bL-\bL^*-\bS^*]_{i,\cdot}^{(\gamma n_2)}|\leq |[\bL-\bL^*]_{i,\cdot}^{((\gamma-\gamma^*) n_2)}|,
\]
we have 
\begin{align*}
|\bD_{ij}| =& | [\bL-\bL^*-F(\bL-\bL^*-\bS^*)]_{ij}|\\\leq& |[\bL-\bL^*]_{ij}|+\max(|[\bL-\bL^*-\bS^*]_{i,\cdot}^{(\gamma n_2)}|, |[\bL-\bL^*-\bS^*]_{\cdot,j}^{(\gamma n_1)}|)\\
 \leq & |[\bL-\bL^*]_{ij}|+\max(|[\bL-\bL^*]_{i,\cdot}^{((\gamma-\gamma^*) n_2)}|, |[\bL-\bL^*]_{\cdot,j}^{((\gamma-\gamma^*) n_1)}|).
\end{align*}

Applying the estimations above, and repeatedly use the fact that $(x+y)^2\leq 2x^2+2y^2$, we have
\begin{align}\nonumber
&\|\bD\|_F^2=\sum_{(i,j)\in\mathcal{S}}\bD_{ij}^2\leq \sum_{ij\in \mathcal{S}_1}[\bL-\bL^*]_{ij}^2
\\\nonumber+&\sum_{ij\in \mathcal{S}_2}\left\{|[\bL-\bL^*]_{ij}|+\max(|[\bL-\bL^*]_{i,\cdot}^{((\gamma-\gamma^*) n_2)}|, |[\bL-\bL^*]_{\cdot,j}^{((\gamma-\gamma^*) n_1)}|)\right\}^2\\\nonumber
\leq & \sum_{ij\in \mathcal{S}_1}[\bL-\bL^*]_{ij}^2+2\sum_{ij\in \mathcal{S}_2}[\bL-\bL^*]_{ij}^2+2\sum_{ij\in \mathcal{S}_2}\max\{|[\bL-\bL^*]_{i,\cdot}^{((\gamma-\gamma^*) n_2)}|, |[\bL-\bL^*]_{\cdot,j}^{((\gamma-\gamma^*) n_1)}|\}^2\\\nonumber
\leq & \sum_{ij\in \mathcal{S}_1}[\bL-\bL^*]_{ij}^2+2\sum_{ij\in \mathcal{S}_2}[\bL-\bL^*]_{ij}^2+2\sum_{ij\in \mathcal{S}_2}\{|[\bL-\bL^*]_{i,\cdot}^{((\gamma-\gamma^*) n_2)}|^2 + |[\bL-\bL^*]_{\cdot,j}^{((\gamma-\gamma^*) n_1)}|^2\}\\\nonumber
\leq & \sum_{ij\in \mathcal{S}}[\bL-\bL^*]_{ij}^2+\sum_{ij\in \mathcal{S}_2}[\bL-\bL^*]_{ij}^2+4\frac{\gamma^*}{\gamma-\gamma^*}\|\bL-\bL^*\|_F^2\\\nonumber
\leq & 2\sum_{ij\in \mathcal{S}}[\bP_{\mathcal{T}_{\bL^*}}(\bL-\bL^*)]_{ij}^2+2\sum_{ij\in \mathcal{S}_2}[\bP_{\mathcal{T}_{\bL^*}}(\bL-\bL^*)]_{ij}^2+4\frac{\gamma^*}{\gamma-\gamma^*}\|\bL-\bL^*\|_F^2 \\\nonumber+&
2\sum_{ij\in \mathcal{S}}[\bL-\bL^*-\bP_{\mathcal{T}_{\bL^*}}(\bL-\bL^*)]_{ij}^2+2\sum_{ij\in \mathcal{S}_2}[\bL-\bL^*-\bP_{\mathcal{T}_{\bL^*}}(\bL-\bL^*)]_{ij}^2
\\\nonumber\leq & 2\sum_{ij\in \mathcal{S}}[\bP_{\mathcal{T}_{\bL^*}}(\bL-\bL^*)]_{ij}^2+2\sum_{ij\in \mathcal{S}_2}[\bP_{\mathcal{T}_{\bL^*}}(\bL-\bL^*)]_{ij}^2+4\frac{\gamma^*}{\gamma-\gamma^*}\|\bL-\bL^*\|_F^2 \\&+ 4 \|\bL-\bL^*-\bP_{\mathcal{T}_{\bL^*}}(\bL-\bL^*)\|_F^2.\label{eq:estimation1}
\end{align}

In another aspect, Lemma~\ref{lemma:approximation} implies
\begin{equation}
\|\bL-\bL^*-\bP_{\mathcal{T}_{\bL^*}}(\bL-\bL^*)\|_F\leq \frac{a}{2}\|\bL-\bL^*\|_F.\label{eq:estimation2}
\end{equation}
In addition, since there exists $\bA\in\reals^{n_2\times r},\bB\in\reals^{n_1\times r}$ such that $\bP_{\mathcal{T}_{\bL^*}}(\bL-\bL^*)=\bU\bA^T+\bB\bV^T$, and  for each row or column, at most $\gamma+\gamma^*$ percentage of points lie in $\mathcal{S}$, 
\begin{align}\nonumber
&\sum_{(i,j)\in\mathcal{S}}[\bP_{\mathcal{T}_{\bL^*}}(\bL-\bL^*)]_{ij}^2\leq 2 \sum_{(i,j)\in\mathcal{S}}[\|(\bU\bA)_{ij}\|^2+\|(\bB\bV^T)_{ij}\|^2]\\\leq&  2  (\gamma+\gamma^*)\mu r \sum_{1\leq i\leq n_1, 1\leq j\leq n_2}[\|(\bU\bA)_{ij}\|^2+\|(\bB\bV^T)_{ij}\|^2]\leq 2  (\gamma+\gamma^*)\mu r  \|\bP_{\mathcal{T}_{\bL^*}}(\bL-\bL^*)\|_F^2\nonumber\\
\leq &2(\gamma+\gamma^*)\mu r \|\bL-\bL^*\|_F^2.
\label{eq:estimation3}
\end{align}
Similarly,
\begin{equation}\label{eq:estimation4}
\sum_{(i,j)\in\mathcal{S}_2}[\bP_{\mathcal{T}_{\bL^*}}(\bL-\bL^*)]_{ij}^2\leq 2 \gamma^*\mu r \|\bL-\bL^*\|_F^2,
\end{equation}

Combining \eqref{eq:estimation1}-\eqref{eq:estimation4}, \eqref{eq:second_bound} is proved. 
\end{proof}

\paragraph*{Proof of Lemma~\ref{lemma:bD}(b)}
\begin{proof}
Let $\bL'=\bL-\bN^*$, then applying the fact that for any $\bx,\by\in\reals^n$,
\[
|[\bx+\by]^{(k)}|\leq |[\bx]^{(k)}|+|[\bx]^{(1)}|,
\]
we have
\begin{align*}
\|\bD'\|_{F}^{2} & \leq\sum_{ij\in\mathcal{S}}[\bL'-\bL^{*}]_{ij}^{2}+\sum_{ij\in\mathcal{S}_{2}}[\bL'-\bL^{*}]_{ij}^{2}\\
 & +2\sum_{ij\in\mathcal{S}_{2}}\{([\bL'-\bL^{*}]_{i,\cdot}^{((\gamma-\gamma^{*})n_{2})})^{2}+([\bL'-\bL^{*}]_{\cdot,j}^{((\gamma-\gamma^{*})n_{1})})^{2}\}\\
 & \leq2\left(\sum_{ij\in\mathcal{S}}[\bL-\bL^{*}]_{ij}^{2}+[\bN_{ij}^{*}]^{2}+\sum_{ij\in\mathcal{S}_{2}}[\bL-\bL^{*}]_{ij}^{2}+[\bN_{ij}^{*}]^{2}\right)\\
 & +4\sum_{ij\in\mathcal{S}_{2}}\{([\bL-\bL^{*}]_{i,\cdot}^{((\gamma-\gamma^{*})n_{2})})^{2}+[\bN^{*}]_{i,\cdot}^{(1)^{2}}+([\bL-\bL^{*}]_{\cdot,j}^{((\gamma-\gamma^{*})n_{1})})^{2}+[\bN^{*}]_{\cdot,j}^{(1)^{2}}\}\\
 & \leq2C_{1}^{2}\|\bL-\bL^{*}\|_{F}^{2}+2(\gamma+5\gamma^{*})N_{1},
\end{align*}
where the last inequality follows from the proof of part (a) and the definition of $N_1$.\end{proof}

\paragraph*{Proof of Lemma~\ref{lemma:bD2}}
\begin{proof}
Following \eqref{eq:estimation1} and the proof of Lemma~\ref{lemma:bD}[a], and note that Lemma~\ref{lemma:sampling} means that $\gamma^*$ and $\gamma$ are replaced by arbitrary numbers in the intervals $[0.5 p \gamma^*, 1.5 p \gamma^*]$ and $[0.5p\gamma, 1.5p\gamma]$, we have
\begin{align*}
\|\tilde{\bD}\|_F^2\leq 6(\gamma+2\gamma^*)p\mu r \|\bL-\bL^*\|_F^2+4\frac{3\gamma^*}{\gamma-3\gamma^*}\|P_{\mathbf{\Phi}}(\bL-\bL^*)\|_F^2+a^2\|\bL-\bL^*\|_F^2.
\end{align*}
Applying Lemma~\ref{lemma:approximation} and \eqref{eq:estimation2}, we have
\begin{align*}
\|P_{\mathbf{\Phi}}(\bL-\bL^{*})\|_{F} & \leq\|P_{\mathbf{\Phi}}P_{T_{\bL^{*}}}(\bL-\bL^{*})\|_{F}+\|P_{\mathbf{\Phi}}P_{T_{\bL^{*}}}^{\perp}(\bL-\bL^{*})\|_{F}\\
 & \leq\sqrt{p(1+\epsilon)}\|\bL-\bL^{*}\|_{F}+\frac{a}{2}\|\bL-\bL^{*}\|_{F}.
\end{align*}
Combining it with the estimation of $\|P_{\mathbf{\Phi}}(\bL-\bL^*)\|_F$ in Lemma~\ref{lemma:approximation}, we have $\|\tilde{\bD}\|_F\leq \tilde{C}_1\|P_{\mathbf{\Phi}}(\bL-\bL^*)\|_F$ with
\[
\tilde{C}_1=\frac{1}{p(1-\epsilon)}\Big[6(\gamma+2\gamma^*)p\mu r+4\frac{3\gamma^*}{\gamma-3\gamma^*}(\sqrt{p(1+\epsilon)}+\frac{a}{2})^2+a^2\Big].
\]
\end{proof}

\paragraph*{Proof of Lemma~\ref{lemma:approximation}}
\begin{proof}
Let the SVD decomposition of $\bL^*$ be $\bL^*=\bU\bSigma\bV$,   $\bU^\perp$ and $\bV^\perp$ be orthogonal matrices of sizes $\reals^{n_1\times (n_1-r)}$ and $\reals^{n_2\times (n_2-r)}$ such that $\Col(\bU^\perp)\perp\Col(\bU)$ and $\Col(\bV^\perp)\perp\Col(\bV)$ (here $\Col(\bU)$ represents the spanned  spanned by the columns of $\bU$), and
\[
\bL^*_{(1,1)}=\bU^T\bL^*\bV, \bL^*_{(1,2)}=\bU^T\bL^*\bV^\perp, \bL^*_{(2,1)}=\bU^{\perp T}\bL^*\bV, \bL^*_{(2,2)}=\bU^{\perp T}\bL^*\bV^\perp.
\]
Since $\rank(\bL^*)=r$, we have
\[
\bL^*_{(2,2)}=\bL^*_{(2,1)}{\bL^*_{(1,1)}}^{-1}\bL^*_{(1,2)}.
\]
Since all singular values of $\bL^*_{(1,1)}$ are larger than $(1-a)\sigma_{r}(\bL^*)$, if the singular value decomposition of ${\bL^*_{(1,1)}}^{-1}$ is given by \[
{\bL^*_{(1,1)}}^{-1}=\bU_0\bSigma_0\bV_0^T,
\]
then the $\|\bSigma_0\|\leq 1/(1-a)\sigma_{r}(\bL^*)$. Applying 
\[
\|\bA\bB\|_F^2\leq \|\bA\|_F^2\|\bB\|_F^2
\]
and the fact that for a square, diagonal matrix $\Sigma$, $|[\bX\bSigma]_{ij}|=|\bX_{ij}\bSigma_{jj}|\leq \|\bSigma\||\bX_{ij}|$, we have 

\begin{align}
\|\bL_{(2,2)}^{*}\|_{F} & =\|\bL_{(2,1)}^{*}\bU_{0}\bSigma_{0}\bV_{0}^{T}\bL_{(1,2)}^{*}\|_{F}\nonumber\\
 & \leq\|\bL_{(2,1)}^{*}\bU_{0}\bSigma_{0}\|_{F}\|\bV_{0}^{T}\bL_{(1,2)}^{*}\|_{F}\nonumber\\
 & \leq\frac{1}{(1-a)\sigma_{r}(\bL^{*})}\|\bL_{(2,1)}^{*}\bU_{0}\|_{F}\|\bV_{0}^{T}\bL_{(1,2)}^{*}\|_{F}\nonumber\\
 & \leq\frac{1}{(1-a)\sigma_{r}(\bL^{*})}\|\bL_{(2,1)}^{*}\|_{F}\|\bL_{(1,2)}^{*}\|_{F}\nonumber\\
 & \leq\frac{1}{(1-a)\sigma_{r}(\bL^{*})}\Big(\frac{\|\bL_{(2,1)}^{*}\|_{F}^{2}+\|\bL_{(1,2)}^{*}\|_{F}^{2}}{2}\Big)\nonumber\\
 & \leq\frac{1}{(1-a)\sigma_{r}(\bL^{*})}\Big(\frac{a^{2}\sigma_{r}(\bL^{*})^{2}}{2}\Big)\nonumber\\
 & \leq\frac{a^{2}}{2(1-a)}\sigma_{r}(\bL^{*}),\label{eq:temp}
\end{align}

and \eqref{eq:control1} is proved. The proof of \eqref{eq:control2} is similar.
\end{proof}

\paragraph*{Proof of Lemma~\ref{lemma:retraction}}
\begin{proof}
Let the SVD decomposition of $\bL$ be $\bL=\bU\bSigma\bV$,  and
\begin{align*}
\bL_{(1,1)} & =\bU^{T}(\bX+\bL)\bV,\bL_{(1,2)}=\bU^{T}(\bX+\bL)\bV^{\perp}\\
 & =\bU^{T}\bX\bV^{\perp},\bL_{(2,1)}=\bU^{\perp T}(\bX+\bL)\bV=\bU^{\perp T}\bX\bV,
\end{align*}
then it is clear that
\[
R^{(2)}_{\bL}(\bX)=\bL+\bX+\bU^{\perp}\bL_{(2,1)}{\bL_{(1,1)}}^{-1}\bL_{(1,2)}\bV^{\perp T},
\]
and using the same argument as in \eqref{eq:temp}, 
\begin{align*}
\|\bL_{(2,1)}{\bL_{(1,1)}}^{-1}\bL_{(1,2)}\|_{F}\leq & \frac{1}{\sigma_{r}(\bL_{(1,1)}^{*})}\|\bL_{(1,2)}\|_{F}\|\bL_{(2,1)}\|_{F}\\
\leq & \frac{1}{\sigma_{r}(\bL)-\|\bX\|}\Big(\frac{\|\bL_{(2,1)}\|_{F}^{2}+\|\bL_{(1,2)}\|_{F}^{2}}{2}\Big)\\
\leq & \frac{1}{\sigma_{r}(\bL)-\|\bX\|}\frac{\|\bX\|_{F}^{2}}{2}.
\end{align*}
So Lemma~\ref{lemma:retraction} is proved for $R^{(2)}_{\bL}(\bX)$.

By definition, $R^{(1)}_{\bL}(\bX)$ is the closest matrix to $\bL+\bX$ that has rank $r$, so $\|R^{(1)}_{\bL}(\bX)-(\bL+\bX)\|_F\leq R^{(2)}_{\bL}(\bX)-(\bL+\bX)\|_F$ and Lemma~\ref{lemma:retraction} is also proved for $R^{(1)}_{\bL}(\bX)$.
\end{proof}

\paragraph*{Proof of Lemma~\ref{lemma:prob}}
\begin{proof}
In the proof WLOG we assume $\sigma=1$ and the generic cases can be proved similarly.\\
 (a) It follows from the estimation of distribution of the maximum of $n_1$ i.i.d. Gaussian variables $\{g_i\}_{i=1}^{n_1}$:
\begin{align*}
&\Pr\{\max_{1\leq i\leq n_1}|g_i|\leq 4\sqrt{\ln (n_1n_2)}\}\geq \Big(1-2\exp\Big(-\frac{(4\sqrt{\ln (n_1n_2)})^2}{2}\Big)\Big)^{n_1}\\\geq& 1-2n_1\exp\Big(-\frac{(4\sqrt{\ln (n_1n_2)})^2}{2}\Big)=1-2n_1^{-7}n_2^{-8},
\end{align*}
where the first inequality applies the estimation of the cumulative distribution function of the Gaussian distribution~\cite[pg 8]{ledoux1991probability}.

Combining this estimation for each column of $\bN^*$ and applying the union bound, the second inequality in part (a) holds with probability $1-2n_1^{-7}n_2^{-7}$. Similarly, the first inequality in part (a) holds with the same probability.

(b) First, we parameterize $\bL$ by $g(\bL)=P_{\bL^*}(\bL-\bL^*)$. 
Then we claim that, for any $\bL$ and $\bL'$ such that $\|\bL-\bL^*\|_F, \|\bL'-\bL^*\|_F\leq a\sigma_r(\bL^*)$, there exists $C_0$ depending on $a$ such that
\begin{equation}\label{eq:claim}
\|P_{T(\bL)}(\bL-\bL^*)-P_{T(\bL')}(\bL'-\bL^*)\|_F\leq C_0 \|g(\bL)-g(\bL')\|_F.
\end{equation}
To prove \eqref{eq:claim}, apply \eqref{eq:control2} and obtain
\begin{equation}\label{eq:intermediate1}
\|\bL-\bL'\|_F\leq \frac{1}{1-\frac{a}{2}}\|g(\bL)-g(\bL')\|_F.
\end{equation}

Since $P_{T(\bL)}=\bU_{\bL}\bU_{\bL}^T+\bV_{\bL}\bV_{\bL}^T-\bU_{\bL}\bU_{\bL}^T\bV_{\bL}\bV_{\bL}^T$, and using Davis-Kahan theorem~\cite{doi:10.1137/0707001} and the assumption $\|\bL-\bL^*\|_F\leq a\sigma_r(\bL^*)$, there exists $c_1$, $c_2$ depending on $a$ such that  
\[
\|\bU_{\bL}\bU_{\bL}^T-\bU_{\bL'}\bU_{\bL'}^T\|_F\leq c_1,\,\,
\|\bV_{\bL}\bV_{\bL}^T-\bV_{\bL'}\bV_{\bL'}^T\|_F\leq c_2,
\]
so there exists $C'$ depending on $a$ such that
\begin{align}\label{eq:intermediate2}
&\|P_{T(\bL)}(\bL-\bL^*)-P_{T(\bL')}(\bL'-\bL^*)\|_F\\ \nonumber=&\|[P_{T(\bL')}(\bL-\bL^*)-P_{T(\bL')}(\bL'-\bL^*)]+[P_{T(\bL)}(\bL-\bL^*)-P_{T(\bL')}(\bL-\bL^*)]\|_F\\\leq&\|\bL-\bL'\|_F+C'\|\bL-\bL'\|_F. \nonumber
\end{align}
Combining \eqref{eq:intermediate1} and \eqref{eq:intermediate2},  \eqref{eq:claim} is proved.

Second, based on \eqref{eq:claim}, we will apply an $\epsilon$-net covering argument to finish the proof that combines probabilistic estimation for each $\bL$ and a union bound ($\epsilon$-net covering argument is a standard argument in probabilistic estimation \cite{vershynin12}). Use the estimation of the cumulative distribution function of the Gaussian distribution~\cite[pg 8]{ledoux1991probability}, for any $\bL'$,
\[
\Pr\left\{\langle\bN^*, P_{T_{\bL'}} (\bL'-\bL^*)\rangle\geq t\|P_{T_{\bL'}} (\bL'-\bL^*)\|_F\right\}\leq \frac{1}{2}\exp\left(-\frac{t^2}{2}\right).
\]
For any $\bL$ such that $\|g(\bL')-g(\bL)\|_F<\epsilon$, applying \eqref{eq:claim},
\[
\Pr\left\{\langle\bN^*, P_{T_{\bL}} (\bL-\bL^*)\rangle\geq t\|P_{T_{\bL}} (\bL-\bL^*)\|_F+C_0\epsilon(\|\bN^*\|_F+t) \right\}\leq \frac{1}{2}\exp\left(-\frac{t^2}{2}\right).
\]
Using union bound, there is an $\epsilon$-net of the set $\{g(\bL):\|g(\bL)\|_F=x\}$ with at most $(C_5x/\epsilon)^{n_1r+n_2r-r^2}$ points.  Therefore, for all $\bL$ such that $x-\epsilon\leq \|P_{T_{\bL}} (\bL-\bL^*)\|_F\leq x+\epsilon$, 
\begin{align}
 & \Pr\left\{ \langle\bN^{*},P_{T_{\bL}}(\bL-\bL^{*})\rangle\geq t\|P_{T_{\bL}}(\bL-\bL^{*})\|_{F}+2C_{0}\epsilon(\|\bN^{*}\|_{F}+t)\right\} \nonumber\\
\leq & \frac{1}{2}\exp\left(-\frac{t^{2}}{2}\right)\cdot\Big(\frac{C_{5}x}{\epsilon}\Big)^{n_{1}r+n_{2}r-r^{2}}.\label{eq:covering1}
\end{align}
Let $t=x/8$ and $\epsilon=x/16C_0\|\bN^*\|_F$, then when $\|\bN^*\|_F\geq 1$ (which holds with high probability as $n_1n_2$ goes to infinity), then using $C_0\geq 1$ we have $\epsilon\leq x/16$, and when $x\geq 4$,
\begin{align}\nonumber
&t\|P_{T_{\bL}} (\bL-\bL^*)\|_F+2C_0\epsilon(\|\bN^*\|_F+t) 
\leq \frac{x}{8}(x+\epsilon)+\frac{x}{8\|\bN^*\|_F}(\|\bN^*\|_F+t)
\\\nonumber=& \frac{x}{8}(x+\epsilon)+\frac{x}{8}+\frac{x^2}{64\|\bN^*\|_F}\nonumber
\leq  \frac{x^2}{8}\frac{17}{16}+\frac{x}{8}+\frac{x^2}{64}\leq \frac{x^2}{8}\frac{17}{16}+\frac{x^2}{32}+\frac{x^2}{64} 
\leq  \frac{1}{4}(x-\epsilon)^2\\\leq&  \frac{1}{4}\|P_{T_{\bL}} (\bL-\bL^*)\|_F^2,\label{eq:covering2}
\end{align}
where the last inequality applies the assumption  $x-\epsilon\leq \|P_{T_{\bL}} (\bL-\bL^*)\|_F$.
Combining \eqref{eq:covering1} and \eqref{eq:covering2} and recall that $t=x/8$, we have that for all $\bL$ such that $x-x/16C_0\|\bN^*\|_F\leq \|P_{T_{\bL}} (\bL-\bL^*)\|_F\leq x+x/16C_0\|\bN^*\|_F$, 
\begin{align}
\Pr\Biggl\{ & \langle\bN^{*},P_{T_{\bL}}(\bL-\bL^{*})\rangle\geq\frac{1}{4}\|P_{T_{\bL}}(\bL-\bL^{*})\|_{F}^{2},\nonumber\\
 & \text{for all \ensuremath{\bL} s.t. \ensuremath{\Big|\|P_{T_{\bL}}(\bL-\bL^{*})\|_{F}-x\Big|\leq\frac{x}{16C_{0}\|\bN^{*}\|_{F}}}}\Biggl\}\nonumber\\
\leq & \frac{1}{2}\exp\left(-\frac{x^{2}}{128}\right)\cdot\Big(16C_{5}C_{0}\|\bN^{*}\|_{F}\Big)^{n_{1}r+n_{2}r-r^{2}}.\label{eq:covering3}
\end{align}
Let $x_i=\sqrt{n_1+n_2+128(n_1r+n_2r-r^2)\ln(16C_5C_0\|\bN^*\|_F)} (1+1/16C_0\|\bN^*\|_F)^i$ with $i=1,2,...$, then
\begin{align}
 & \sum_{i=1}^{\infty}\exp\left(-\frac{x_{i}^{2}}{128}\right)\cdot\Big(16C_{5}C_{0}\|\bN^{*}\|_{F}\Big)^{n_{1}r+n_{2}r-r^{2}}\nonumber\\
\leq & \exp(-\frac{n_{1}+n_{2}}{128})\sum_{i=1}^{\infty}\exp(-{(1+1/16C_{0}\|\bN^{*}\|_{F})^{2i}})\nonumber\\
\leq & \exp(-\frac{n_{1}+n_{2}}{128})\sum_{i=1}^{\infty}\exp(-1-i/8C_{0}\|\bN^{*}\|_{F})\nonumber\\
= & \exp(-\frac{n_{1}+n_{2}}{128}-1)\frac{\exp(-1/8C_{0}\|\bN^{*}\|_{F})}{1-\exp(-1/8C_{0}\|\bN^{*}\|_{F})}\nonumber\\
\leq & 8C_{0}\|\bN^{*}\|_{F}\exp(-\frac{n_{1}+n_{2}}{128}-1),\label{eq:convering4}
\end{align}
where the last inequality uses $\exp(-c)\leq 1-c$ when $c\geq 0$. Clearly, the RHS goes to $0$ as $n_1+n_2\rightarrow\infty$.

Combining the estimation \eqref{eq:covering3} for $\{x_i\}_{i=1}^\infty$, with probability $1-8C_0\|\bN^*\|_F\exp(-\frac{n_1+n_2}{128}-1)$, the event \eqref{eq:event_noisy} holds for all $\bL$ such that \[\|g(\bL)\|_F\geq \max(\sqrt{n_1+n_2+128(n_1r+n_2r-r^2)\ln(16C_5C_0\|\bN^*\|_F)},4).\] Combining it with \eqref{eq:control2}, the event \eqref{eq:event_noisy} holds for all for all $\bL$ such that
\begin{align*}
a\sigma_{r}(\bL^{*}) & \geq\|\bL-\bL^{*}\|_{F}\\
 & \geq\frac{1}{1-\frac{a}{2}}\max(\sqrt{n_{1}+n_{2}+128(n_{1}r+n_{2}r-r^{2})\ln(16C_{5}C_{0}\|\bN^{*}\|_{F})},4).
\end{align*}
Considering that $\sqrt{n_1+n_2+128(n_1r+n_2r-r^2)\ln(16C_5C_0\|\bN^*\|_F)}$ is the dominant term when $n_1,n_2\rightarrow\infty$, Lemma~\ref{lemma:prob}(b) is proved.
\end{proof}
\bibliographystyle{abbrv}
\bibliography{bib-online}

\begin{thebibliography}{10}

\bibitem{absil2009optimization}
P.~Absil, R.~Mahony, and R.~Sepulchre.
\newblock {\em Optimization Algorithms on Matrix Manifolds}.
\newblock Princeton University Press, 2009.

\bibitem{Absil2015}
P.-A. Absil and I.~V. Oseledets.
\newblock {Low-rank retractions: a survey and new results}.
\newblock {\em Computational Optimization and Applications}, 62(1):5--29, 2015.

\bibitem{pmlr-v49-bandeira16}
A.~S. Bandeira, N.~Boumal, and V.~Voroninski.
\newblock On the low-rank approach for semidefinite programs arising in
  synchronization and community detection.
\newblock In V.~Feldman, A.~Rakhlin, and O.~Shamir, editors, {\em 29th Annual
  Conference on Learning Theory}, volume~49 of {\em Proceedings of Machine
  Learning Research}, pages 361--382, Columbia University, New York, New York,
  USA, 23--26 Jun 2016. PMLR.

\bibitem{Basri03}
R.~Basri and D.~Jacobs.
\newblock Lambertian reflectance and linear subspaces.
\newblock {\em IEEE Transactions on Pattern Analysis and Machine Intelligence},
  25(2):218--233, February 2003.

\bibitem{Bhojanapalli:2015:TLA:2722129.2722191}
S.~Bhojanapalli, P.~Jain, and S.~Sanghavi.
\newblock Tighter low-rank approximation via sampling the leveraged element.
\newblock In {\em Proceedings of the Twenty-sixth Annual ACM-SIAM Symposium on
  Discrete Algorithms}, SODA '15, pages 902--920, Philadelphia, PA, USA, 2015.
  Society for Industrial and Applied Mathematics.

\bibitem{doi:10.1137/16M105808X}
N.~Boumal.
\newblock Nonconvex phase synchronization.
\newblock {\em SIAM Journal on Optimization}, 26(4):2355--2377, 2016.

\bibitem{Burer2003}
S.~Burer and R.~D. Monteiro.
\newblock A nonlinear programming algorithm for solving semidefinite programs
  via low-rank factorization.
\newblock {\em Mathematical Programming}, 95(2):329--357, 2003.

\bibitem{Burer2005}
S.~Burer and R.~D. Monteiro.
\newblock Local minima and convergence in low-rank semidefinite programming.
\newblock {\em Mathematical Programming}, 103(3):427--444, 2005.

\bibitem{robust_pca09}
E.~J. Cand\`{e}s, X.~Li, Y.~Ma, and J.~Wright.
\newblock Robust principal component analysis?
\newblock {\em J. ACM}, 58(3):11:1--11:37, June 2011.

\bibitem{Chandrasekaran_Sanghavi_Parrilo_Willsky_2009}
V.~Chandrasekaran, S.~Sanghavi, P.~A. Parrilo, and A.~S. Willsky.
\newblock Rank-sparsity incoherence for matrix decomposition.
\newblock {\em SIAM Journal on Optimization}, 21(2):572--596, 2011.

\bibitem{ChenWainwright2015}
Y.~Chen and M.~J. Wainwright.
\newblock Fast low-rank estimation by projected gradient descent: General
  statistical and algorithmic guarantees.
\newblock {\em CoRR}, abs/1509.03025, 2015.

\bibitem{DBLP:journals/corr/CherapanamjeriG16}
Y.~Cherapanamjeri, K.~Gupta, and P.~Jain.
\newblock Nearly-optimal robust matrix completion.
\newblock {\em CoRR}, abs/1606.07315, 2016.

\bibitem{Clarkson:2013:LRA:2488608.2488620}
K.~L. Clarkson and D.~P. Woodruff.
\newblock Low rank approximation and regression in input sparsity time.
\newblock In {\em Proceedings of the Forty-fifth Annual ACM Symposium on Theory
  of Computing}, STOC '13, pages 81--90, New York, NY, USA, 2013. ACM.

\bibitem{doi:10.1137/0707001}
C.~Davis and W.~M. Kahan.
\newblock The rotation of eigenvectors by a perturbation. iii.
\newblock {\em SIAM Journal on Numerical Analysis}, 7(1):1--46, 1970.

\bibitem{DeSa:2015:GCS:3045118.3045366}
C.~De~Sa, K.~Olukotun, and C.~R{\'e}.
\newblock Global convergence of stochastic gradient descent for some non-convex
  matrix problems.
\newblock In {\em Proceedings of the 32Nd International Conference on
  International Conference on Machine Learning - Volume 37}, ICML'15, pages
  2332--2341. JMLR.org, 2015.

\bibitem{ASI:ASI1}
S.~Deerwester, S.~T. Dumais, G.~W. Furnas, T.~K. Landauer, and R.~Harshman.
\newblock Indexing by latent semantic analysis.
\newblock {\em Journal of the American Society for Information Science},
  41(6):391--407, 1990.

\bibitem{Epstein95}
R.~Epstein, P.~Hallinan, and A.~Yuille.
\newblock $5\pm 2$ eigenimages suffice: {A}n empirical investigation of
  low-dimensional lighting models.
\newblock In {\em IEEE Workshop on Physics-based Modeling in Computer Vision},
  pages 108--116, June 1995.

\bibitem{Frieze:2004:FMA:1039488.1039494}
A.~Frieze, R.~Kannan, and S.~Vempala.
\newblock Fast monte-carlo algorithms for finding low-rank approximations.
\newblock {\em J. ACM}, 51(6):1025--1041, Nov. 2004.

\bibitem{conf/aistats/GuWL16}
Q.~Gu, Z.~Wang, and H.~Liu.
\newblock Low-rank and sparse structure pursuit via alternating minimization.
\newblock In A.~Gretton and C.~C. Robert, editors, {\em AISTATS}, volume~51 of
  {\em JMLR Workshop and Conference Proceedings}, pages 600--609. JMLR.org,
  2016.

\bibitem{Ho03}
J.~Ho, M.~Yang, J.~Lim, K.~Lee, and D.~Kriegman.
\newblock Clustering appearances of objects under varying illumination
  conditions.
\newblock In {\em Proceedings of International Conference on Computer Vision
  and Pattern Recognition}, volume~1, pages 11--18, 2003.

\bibitem{5934412}
D.~Hsu, S.~M. Kakade, and T.~Zhang.
\newblock Robust matrix decomposition with sparse corruptions.
\newblock {\em IEEE Transactions on Information Theory}, 57(11):7221--7234, Nov
  2011.

\bibitem{Jain:2013:LMC:2488608.2488693}
P.~Jain, P.~Netrapalli, and S.~Sanghavi.
\newblock Low-rank matrix completion using alternating minimization.
\newblock In {\em Proceedings of the Forty-fifth Annual ACM Symposium on Theory
  of Computing}, STOC '13, pages 665--674, New York, NY, USA, 2013. ACM.

\bibitem{6319655}
A.~Kyrillidis and V.~Cevher.
\newblock Matrix alps: Accelerated low rank and sparse matrix reconstruction.
\newblock In {\em 2012 IEEE Statistical Signal Processing Workshop (SSP)},
  pages 185--188, Aug 2012.

\bibitem{ledoux1991probability}
M.~Ledoux and M.~Talagrand.
\newblock {\em Probability in Banach Spaces: Isoperimetry and Processes}.
\newblock A Series of Modern Surveys in Mathematics Series. Springer, 1991.

\bibitem{Li_backgroundsubtraction}
L.~Li, W.~Huang, I.~Gu, and Q.~Tian.
\newblock Statistical modeling of complex backgrounds for foreground object
  detection.
\newblock {\em Image Processing, IEEE Transactions on}, 13(11):1459 --1472,
  nov. 2004.

\bibitem{7035075}
X.~Li and J.~Haupt.
\newblock Identifying outliers in large matrices via randomized adaptive
  compressive sampling.
\newblock {\em IEEE Transactions on Signal Processing}, 63(7):1792--1807, April
  2015.

\bibitem{NIPS2011_4486}
L.~W. Mackey, M.~I. Jordan, and A.~Talwalkar.
\newblock Divide-and-conquer matrix factorization.
\newblock In J.~Shawe-Taylor, R.~S. Zemel, P.~L. Bartlett, F.~Pereira, and
  K.~Q. Weinberger, editors, {\em Advances in Neural Information Processing
  Systems 24}, pages 1134--1142. Curran Associates, Inc., 2011.

\bibitem{NIPS2014_5430}
P.~Netrapalli, N.~U~N, S.~Sanghavi, A.~Anandkumar, and P.~Jain.
\newblock Non-convex robust pca.
\newblock In Z.~Ghahramani, M.~Welling, C.~Cortes, N.~D. Lawrence, and K.~Q.
  Weinberger, editors, {\em Advances in Neural Information Processing Systems
  27}, pages 1107--1115. Curran Associates, Inc., 2014.

\bibitem{Park2016}
D.~Park, A.~Kyrillidis, S.~Bhojanapalli, C.~Caramanis, and S.~Sanghavi.
\newblock {Provable Burer-Monteiro factorization for a class of
  norm-constrained matrix problems}.
\newblock jun 2016.

\bibitem{pmlr-v54-park17a}
D.~Park, A.~Kyrillidis, C.~Carmanis, and S.~Sanghavi.
\newblock {Non-square matrix sensing without spurious local minima via the
  Burer-Monteiro approach}.
\newblock In A.~Singh and J.~Zhu, editors, {\em Proceedings of the 20th
  International Conference on Artificial Intelligence and Statistics},
  volume~54 of {\em Proceedings of Machine Learning Research}, pages 65--74,
  Fort Lauderdale, FL, USA, 20--22 Apr 2017. PMLR.

\bibitem{Price2006}
A.~L. Price, N.~J. Patterson, R.~M. Plenge, M.~E. Weinblatt, N.~A. Shadick, and
  D.~Reich.
\newblock {Principal components analysis corrects for stratification in
  genome-wide association studies.}
\newblock {\em Nature genetics}, 38(8):904--909, aug 2006.

\bibitem{7809181}
M.~Rahmani and G.~K. Atia.
\newblock High dimensional low rank plus sparse matrix decomposition.
\newblock {\em IEEE Transactions on Signal Processing}, 65(8):2004--2019, April
  2017.

\bibitem{ruhe1974numerical}
A.~Ruhe.
\newblock {\em Numerical Computation of Principal Components when Several
  Observations are Missing}.
\newblock Univ., 1974.

\bibitem{Shalit:2012:OLE:2503308.2188399}
U.~Shalit, D.~Weinshall, and G.~Chechik.
\newblock Online learning in the embedded manifold of low-rank matrices.
\newblock {\em J. Mach. Learn. Res.}, 13(1):429--458, Feb. 2012.

\bibitem{7755794}
J.~Sun, Q.~Qu, and J.~Wright.
\newblock Complete dictionary recovery over the sphere i: Overview and the
  geometric picture.
\newblock {\em IEEE Transactions on Information Theory}, 63(2):853--884, Feb
  2017.

\bibitem{DBLP:conf/icml/TuBSSR16}
S.~Tu, R.~Boczar, M.~Simchowitz, M.~Soltanolkotabi, and B.~Recht.
\newblock Low-rank solutions of linear matrix equations via procrustes flow.
\newblock In {\em Proceedings of the 33nd International Conference on Machine
  Learning, {ICML} 2016, New York City, NY, USA, June 19-24, 2016}, pages
  964--973, 2016.

\bibitem{Vandereycken2013}
B.~Vandereycken.
\newblock Low-rank matrix completion by riemannian optimization.
\newblock {\em SIAM Journal on Optimization}, 23(2):1214--1236, 2013.

\bibitem{vershynin12}
R.~Vershynin.
\newblock Introduction to the non-asymptotic analysis of random matrices.
\newblock In Y.~C. Eldar and G.~Kutyniok, editors, {\em Compressed Sensing:
  Theory and Practice}, pages 210--268. Cambridge University Press, 2012.

\bibitem{pmlr-v54-wang17b}
L.~Wang, X.~Zhang, and Q.~Gu.
\newblock {A Unified Computational and Statistical Framework for Nonconvex
  Low-rank Matrix Estimation}.
\newblock In A.~Singh and J.~Zhu, editors, {\em Proceedings of the 20th
  International Conference on Artificial Intelligence and Statistics},
  volume~54 of {\em Proceedings of Machine Learning Research}, pages 981--990,
  Fort Lauderdale, FL, USA, 20--22 Apr 2017. PMLR.

\bibitem{DBLP:conf/nips/YiPCC16}
X.~Yi, D.~Park, Y.~Chen, and C.~Caramanis.
\newblock Fast algorithms for robust {PCA} via gradient descent.
\newblock In {\em Advances in Neural Information Processing Systems 29: Annual
  Conference on Neural Information Processing Systems 2016, December 5-10,
  2016, Barcelona, Spain}, pages 4152--4160, 2016.

\end{thebibliography}
\end{document}